\documentclass[10pt,twocolumn,letterpaper]{article}

\usepackage[margin=0.75in]{geometry}
\usepackage[T1]{fontenc}
\usepackage{times}
\usepackage{microtype}
\usepackage[hyphens]{url}
\usepackage{graphicx}
\usepackage[numbers,sort&compress]{natbib}
\usepackage[font=small,labelfont=bf]{caption}
\usepackage{booktabs}
\usepackage{multirow}
\usepackage{amsmath}
\usepackage{amssymb}
\usepackage{array}
\usepackage{placeins}
\usepackage{float}
\usepackage{dblfloatfix}
\usepackage{balance}
\usepackage{algorithm}
\usepackage{algpseudocode}
\usepackage{xcolor}
\usepackage[colorlinks=true,allcolors=blue]{hyperref}
\hypersetup{
  pdftitle={StreamSoccer: Event-Driven Memory for Streaming Soccer Commentary},
  pdfauthor={Chenxi Shao, Bozhong Wang, Jiaxin Huang, Zhao Liu, Sunwei Zhu, Tianxin Hang, Gaoqi He, Yang Li, Changbo Wang}
}

\title{StreamSoccer: Event-Driven Memory for Streaming Soccer Commentary}

\author{%
Chenxi Shao\textsuperscript{1,2} \quad
Bozhong Wang\textsuperscript{2,3} \quad
Jiaxin Huang\textsuperscript{2} \quad
Zhao Liu\textsuperscript{2} \quad
Sunwei Zhu\textsuperscript{2}\\
Tianxin Hang\textsuperscript{2} \quad
Gaoqi He\textsuperscript{1} \quad
Yang Li\textsuperscript{1,*} \quad
Changbo Wang\textsuperscript{1,*}\\[0.45em]
\small \textsuperscript{1}East China Normal University, Shanghai, China\\
\small \textsuperscript{2}Migu Video Technology Co., Ltd., Shanghai, China\\
\small \textsuperscript{3}South China University of Technology, Guangzhou, Guangdong, China
}
\date{}

\makeatletter
\g@addto@macro\@maketitle{%
  \begingroup
  \centering
  \vspace{-1.1em}
  \includegraphics[width=\textwidth]{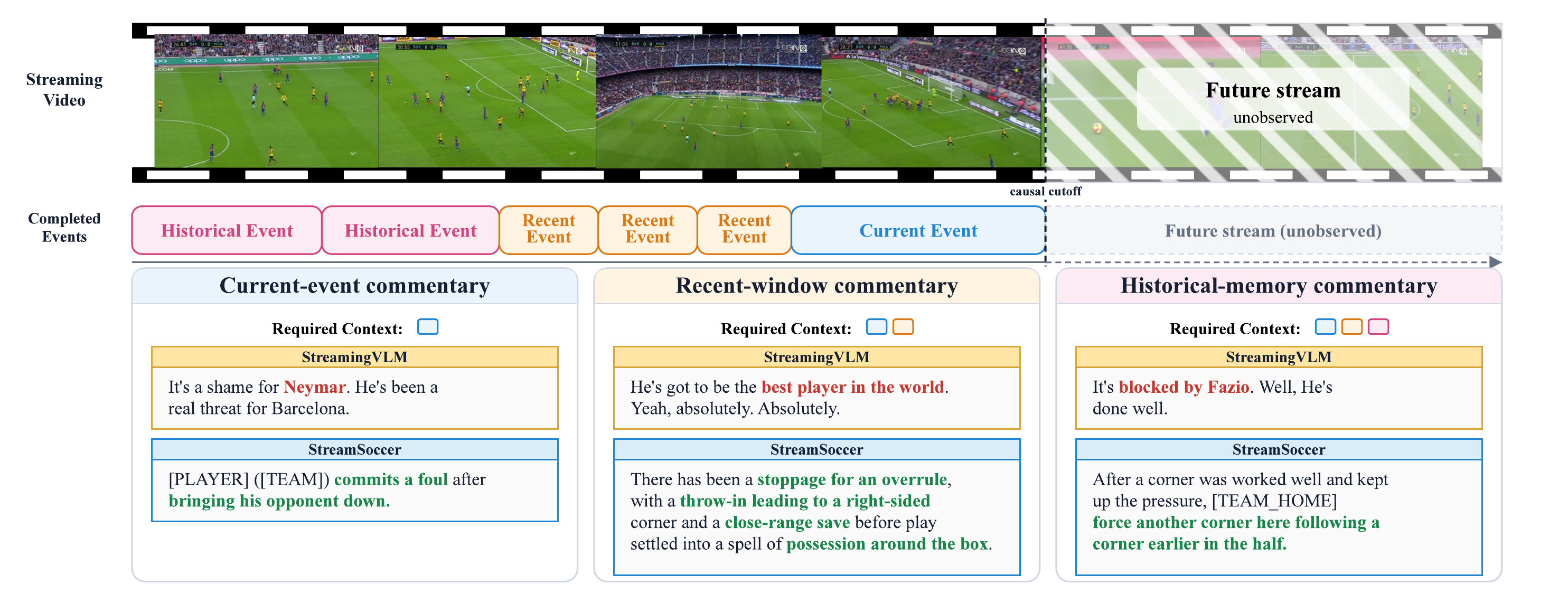}\par
  \captionof{figure}{Completed-event context under a causal stream. The three paired examples compare StreamingVLM and StreamSoccer within one 20-minute match segment selected to contain all three commentary tasks.}
  \label{fig:teaser}
  \endgroup
}
\makeatother

\begin{document}

\maketitle

\begingroup
\renewcommand{\thefootnote}{*}
\footnotetext{Corresponding authors.}
\endgroup
\setcounter{footnote}{0}

\begin{abstract}
Streaming video understanding requires models to causally update state as video
arrives and organize growing history into semantic units that can evolve,
persist, and be recalled under bounded computation and memory. This challenge
is especially pronounced in live soccer commentary, where a system must describe
completed events, summarize recent play, recall earlier events, or remain silent
using only information available before each utterance.
We present StreamSoccer, an event-driven system that uses event memory as its
intermediate representation. A fixed-budget active memory integrates the stream;
completed event states are retained locally and consolidated into retrievable
historical records. A unified generator uses current, recent, and historical
context to produce three commentary modes, while a rule-assisted scheduler
selects a mode or silence. Unlike general streaming video-language models that
organize history around frames, visual tokens, or caches, and soccer-commentary
methods that rely on predefined clips or output timestamps, StreamSoccer
explicitly models event lifecycles.
We construct a three-track streaming soccer commentary dataset and layered
evaluation protocol. At common reference anchors, StreamSoccer obtains CIDEr
scores of 38.62, 23.96, and 17.39 on current-event, recent-window, and
historical-memory commentary, ranking first on the current-event and
historical-memory tracks and second on recent-window. Controlled ablations show
that local completed events improve all tracks; the full system performs best on
all three, with the largest additional gain on recent-window commentary. Across
174 raw-video runs on 58 matches, per-minute RTF p95 remains approximately
0.10--0.22 without sustained growth with match history. These results indicate
that event memory supports streaming soccer commentary across temporal scopes
while controlling long-history computation.
\end{abstract}


\section{Introduction}

Streaming video understanding requires models to update their state continuously
as video arrives and the future remains unobserved, while retaining information
useful for downstream tasks under bounded computation and memory. This challenge
is particularly acute in live soccer commentary: a system must continuously
understand an evolving match process and generate commentary at appropriate
moments across different temporal scopes. Existing soccer-commentary methods
typically rely on predefined video clips, given output timestamps, or offline
full-video input, whereas general streaming video-language models manage growing
history by compressing frames, visual tokens, or key--value caches. These
approaches offer strong video understanding and language generation, but usually
do not treat an evolving semantic process as a state unit that must be maintained
continuously, closed upon completion, and reused later.

This gap concerns not only the efficiency of history compression, but also the
unit around which a streaming system should organize its state. A semantic
process in soccer often spans multiple consecutive video clips: its relevant
visual information accumulates as play unfolds and can continue to influence
subsequent commentary after the process completes. Meanwhile, soccer commentary
does not rely on a single temporal scale. Current-event commentary describes a
just-completed event, recent-window commentary aggregates several recent events,
and historical-memory commentary recalls information formed earlier in the
match. A suitable intermediate representation should therefore absorb new
observations as an event develops, compress a variable-duration process into a
fixed-budget state, form stable and reusable memory upon completion, and support
both recent retention and historical retrieval. Event memory naturally matches
these requirements: a state is activated and updated with an event, closed when
the event completes, and subsequently retained, consolidated, and retrieved when
needed. We therefore hypothesize that event memory is an effective intermediate
representation for streaming soccer commentary.

Based on this hypothesis, we introduce StreamSoccer, an event-driven streaming
soccer commentary system. StreamSoccer maintains a fixed-budget active event
memory as the match progresses and closes the current state as a completed event
memory when an operational event ends. Recent completed events remain compact
latent context, while completed memories can also be consolidated into retrievable
textual records for later historical retrieval. A rule-assisted scheduler
selects current-event commentary, recent-window commentary, historical-memory
commentary, or silence; once a commentary mode is selected, a unified generator
assembles the event context required by that mode and produces the corresponding
commentary. Through this \emph{activate--update--close--consolidate--retrieve}
event lifecycle, StreamSoccer transforms a growing video history into
fixed-budget states with explicit semantic roles.

To train and evaluate these capabilities, we construct three-track streaming
commentary data from SoccerNet action annotations and MatchTime commentary
\citep{deliege2021soccernetv2,rao2024matchtime}. Every sample has an explicit
observation cutoff and source-event provenance; data construction checks for future
information and uses match-disjoint splits. Evaluation separately examines
commentary quality at controlled output anchors, the use of event memory and
historical context, and runtime efficiency from raw-video input.

Our contributions are threefold:

\noindent(1) We formulate soccer commentary as a streaming generation task in
which each output may use only information observable before its emission time.
The task comprises three complementary tracks---current-event, recent-window,
and historical-memory commentary---and is supported by a data pipeline with
explicit observation cutoffs, source provenance, and match-disjoint splits.

\noindent(2) We introduce StreamSoccer, an event-driven streaming soccer
commentary system that uses event memory as its intermediate representation. It
continuously updates fixed-budget active-event state, retains and consolidates
completed event memories, and supports unified commentary generation with recent
context and historical retrieval, thereby organizing a growing video history
into reusable event-level states.

\noindent(3) We evaluate StreamSoccer under distinct settings that separately
measure three-track commentary quality at common reference anchors, the role of
event memory, and long-history processing efficiency from raw-video input.
StreamSoccer ranks first on
current-event and historical-memory commentary and second on recent-window
commentary; the complete event-memory system performs best across all three
tracks in memory-scope ablations, while runtime efficiency shows no sustained
degradation as match history grows.

\begin{figure*}[t]
\centering
\includegraphics[width=0.97\textwidth]{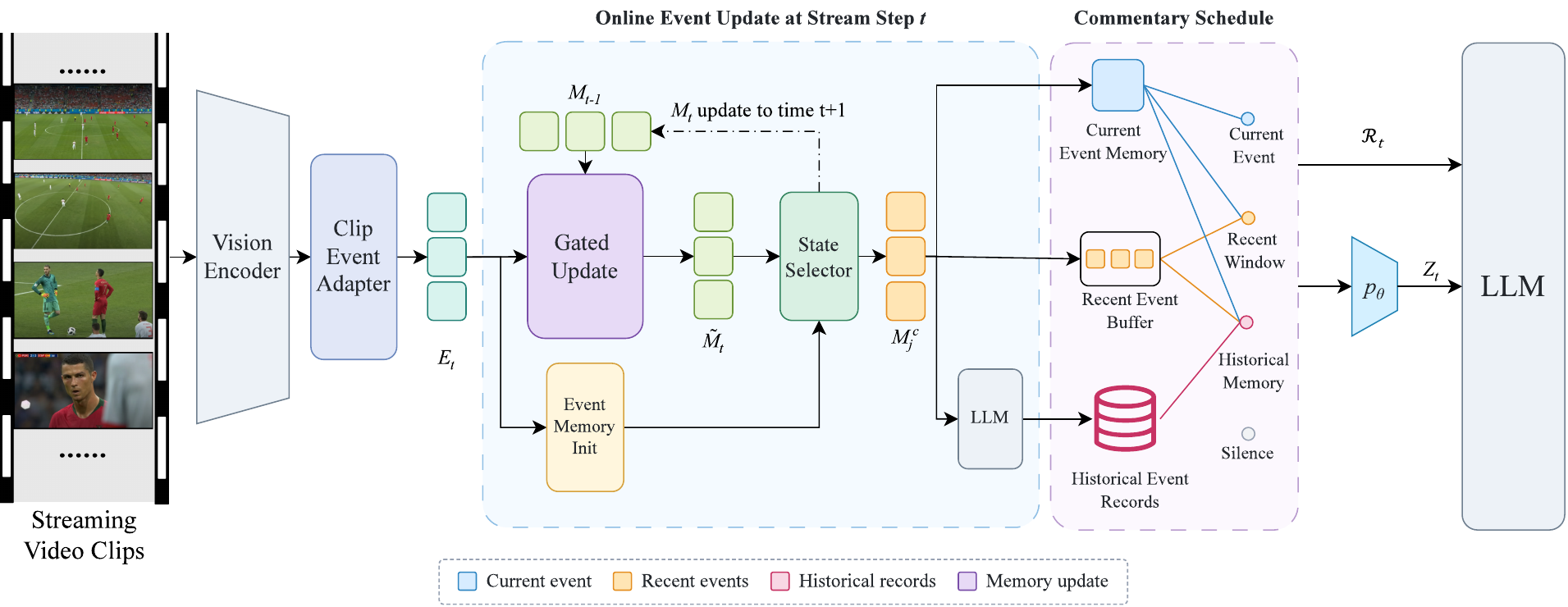}
\caption{Overview of StreamSoccer. Incoming clips are encoded as event tokens
\(E_t\) and update the active event state. When an event closes, its completed
memory \(M_j^{\mathrm c}\) supports current-event commentary, enters the Recent
Event Buffer, and is converted into Historical Event Records. The rule-assisted
schedule selects a commentary mode or silence. For generation, selected latent
states are projected into \(Z_t\), while retrieved records \(\mathcal R_t\) are
provided as text.}
\label{fig:method_overview}
\end{figure*}

\section{Related Work}

\paragraph{Soccer Commentary and Captioning.}
The SoccerNet series established foundational tasks and datasets for soccer video understanding \citep{giancola2018soccernet}. SoccerNet-Caption formulates commentary as timestamped dense video captioning, MatchTime improves video--text alignment, and GOAL provides a knowledge-enhanced commentary benchmark \citep{mkhallati2023soccernetcaption,qi2023goal}. TimeSoccer, \emph{Towards Universal Soccer Video Understanding}, and SoccerMaster further extend full-half modeling and soccer representation learning \citep{you2025timesoccer,rao2025universal,yang2026soccermaster}. MatchAware conditions generation on preceding events, while GameSight combines entity-aware visual reasoning with external statistics and an evolving game state \citep{sun2026matchaware,jin2026gamesight}. These methods provide strong domain data and models but are generally evaluated on predefined clips, reference timestamps, or offline long videos. StreamSoccer instead maintains event state, consolidates completed events, and reuses them while operating along the match timeline.

\paragraph{Streaming Video-Language Models.}
Flash-VStream, VideoStreaming, and StreamFormer support continuous video processing through hierarchical compression, a fixed visual-token budget, and forward-only temporal modeling, respectively \citep{zhang2025flashvstream,qian2024videostreaming,yan2025streamformer}. TimeChat-Online filters redundant visual tokens, while StreamingVLM maintains a compact key--value cache aligned with chunked streaming inference \citep{yao2025timechatonline,xu2026streamingvlm}. VideoLLM-online converts offline temporal annotations into streaming dialogue supervision, while StreamBridge augments offline Video-LLMs with compressed memory and proactive activation \citep{chen2024videollmonline,wang2025streambridge}. These systems demonstrate efficient continuous processing, but usually organize state as task-agnostic frames, tokens, or compressed history rather than around the evolution and lifecycle of soccer events.

\paragraph{Memory and Retrieval for Long Video Understanding.}
MART uses recurrent memory for coherent multi-sentence video description, while MovieChat and MA-LMM compress or write long visual histories into internal memory \citep{lei2020mart,song2024moviechat,he2024malmm}. VideoRAG retrieves external video content, and SoccerComment applies preconstructed multimodal memory and retrieval-augmented prompting to soccer commentary \citep{jeong2025videorag,li2025soccercomment}. Their historical context is typically a generic internal state or a prebuilt corpus. In contrast, StreamSoccer derives retrievable event records from completed memories in the current stream and reuses them as historical context once they become available.

\section{Task Formulation: Streaming Soccer Commentary}
\label{sec:task_formulation}

Streaming soccer commentary is a generation task constrained by temporal
visibility. We represent a match as a temporally ordered sequence of video
clips \(\mathcal C=\{C_1,\ldots,C_T\}\), where the clip \(C_t\) arriving at
step \(t\) covers the interval \([s_t,e_t]\). At decision step \(t\), the
system has an observation cutoff \(\tau_t\) and may access only clips \(C_u\)
satisfying \(e_u\leq\tau_t\), together with auxiliary information formed by
that cutoff.

Based on the match history visible by the cutoff, the system emits one
commentary sentence \(y_t\), or \(\varnothing\) to remain silent. Future video,
events that have not yet occurred, the final match state, and any information
formed after \(\tau_t\) cannot be used for the current output; using them in an
input, conditioning signal, or target constitutes future leakage. This
visibility constraint applies to training samples, controlled evaluation, and
continuous timeline execution.

We define three complementary tracks according to the temporal scope of the
commentary:
\begin{itemize}
\item \textbf{Current-event commentary} describes a just-completed match event,
focusing on its local actions, development, and outcome.
\item \textbf{Recent-window commentary} summarizes multiple recent match
processes visible before the cutoff, capturing short-horizon trends, sequences
of attacks and defenses, or recurring patterns.
\item \textbf{Historical-memory commentary} connects the current or recent
match process with earlier match information that was formed and available
before the cutoff, producing commentary over a longer temporal scope.
\end{itemize}

The three tracks share the same temporal-visibility constraint but differ in
the scope of history they require. This definition does not prescribe a
particular event partition, memory representation, historical retrieval
mechanism, or speaking scheduler.

\section{Method: StreamSoccer}
\label{sec:method}

\subsection{System Overview}

As shown in Figure~\ref{fig:method_overview}, the Vision Encoder and Clip Event
Adapter map each incoming clip to fixed-budget event tokens \(E_t\). Online
Event Update compares the arriving tokens with the previous active memory,
updates the ongoing state through gated integration, and uses the State Selector
to continue that state or initialize a new one. When a reset closes the previous
state, the completed event memory \(M_j^{\mathrm c}\) becomes available for
current-event commentary and is retained in the Recent Event Buffer. Together,
these operations implement the \textbf{Streaming Event Memory Encoder}
described next.

The record branch converts completed event memories into textual Historical
Event Records for later retrieval. At each decision point, the rule-assisted
schedule selects current-event, recent-window, historical-memory commentary, or
silence. For a non-silent mode, the \textbf{Multi-Context Commentary Generator}
projects the required current and recent latent states through \(P_\theta\) into
the soft prefix \(Z_t\); historical-memory commentary additionally supplies the
retrieved records \(\mathcal R_t\) as text to the language model. This separates
compact latent context from textual historical context while retaining one
generation interface across the three commentary modes.

\subsection{Streaming Event Memory Encoder}

For each incoming clip \(C_t\), a frozen Qwen3-VL visual encoder produces
visual tokens.
The Clip Event Adapter compresses them with learnable queries into a fixed
number of clip-level event tokens:
\begin{equation}
V_t=\mathcal F_{\mathrm{vis}}(C_t),
\qquad
E_t=\mathcal A(V_t)\in\mathbb R^{K_m\times d_m}.
\label{eq:clip_event_tokens}
\end{equation}
Here, \(E_t[0]\) summarizes the current clip, while the remaining tokens retain
distributed event information. Before incorporating the clip, the Operational
Transition Head compares \(E_t[0]\) with the previous active state
\(M_{t-1}\) and predicts the operational-transition indicator \(b_t\).

When the clip continues the ongoing process, the memory absorbs incremental
information through slot-wise gating:
\begin{equation}
\widetilde M_t
=(1-G_t)\odot M_{t-1}
+G_t\odot\mathcal U(M_{t-1},E_t).
\label{eq:gated_memory_update}
\end{equation}
Here, \(\mathcal U\) denotes the candidate update after reading the current
clip, and \(G_t\) controls the retention of existing information and the
writing of new information in each memory slot. The state always contains
\(K_m\) tokens, regardless of the number of clips spanned by the process.

In addition to predicted transitions, a maximum event duration \(D_{\max}\)
provides a deterministic safeguard. Let \(\rho_t\) denote this duration
safeguard. The final reset signal and state transition are
\begin{equation}
\begin{aligned}
\rho_t&=[\operatorname{dur}(M_{t-1})\geq D_{\max}],
&
r_t&=b_t\lor\rho_t,\\
M_t&=(1-r_t)\widetilde M_t+r_t\mathcal I(E_t),
&
M_j^{\mathrm c}&\leftarrow M_{t-1}\quad\text{if }r_t=1.
\end{aligned}
\label{eq:event_reset}
\end{equation}
When \(r_t=1\), \(M_{t-1}\) is closed as the immutable completed-event memory
\(M_j^{\mathrm c}\), and the initializer \(\mathcal I\) uses the current clip
\(E_t\) to establish the new \(M_t\). The learned \(b_t\) represents an
operational transition, whereas \(\rho_t\) serves only as a deterministic
duration safeguard. Training uses teacher-forced resets and excludes
duration-safeguard positions from positive transition supervision. At
inference, resets are controlled jointly by the predicted \(b_t\) and
deterministic \(\rho_t\). The remaining active state is closed as the final
completed-event memory at the end of the stream.

\subsection{Memory-to-Record Consolidation}

A completed-event memory \(M_j^{\mathrm c}\) encodes the visual development of
an event in a compact latent form. Memory-to-Record Consolidation maps it through
the memory-to-language projector \(P_\theta(\cdot)\) to a fixed-length soft prefix.
Conditioned on this prefix and the minimal causally available match context
\(x_j^{\mathrm{ctx}}\) at event closure, the language model generates an
event-record caption \(c_j\), followed by deterministic record assembly:
\begin{equation}
\begin{aligned}
p\!\left(c_j\mid P_\theta(M_j^{\mathrm c}),x_j^{\mathrm{ctx}}\right),\\
R_j&=\operatorname{Assemble}\!\left(\operatorname{metadata}_j,c_j\right).
\end{aligned}
\label{eq:memory_to_record}
\end{equation}
The fixed-length prefix keeps the language-model interface independent of the
number of clips within an event. The caption \(c_j\) summarizes the event's
actions, progression, and outcome, while structured metadata and source-event
provenance are populated deterministically. At event closure,
\(M_j^{\mathrm c}\) immediately enters the local event-memory buffer
\(\mathcal B_t\); in parallel, the record branch asynchronously completes
caption generation, deterministic assembly, and storage, after which the ready
\(R_j\) enters \(\mathcal L_t\) for historical retrieval.

\subsection{Multi-Context Commentary Generator}

The Multi-Context Commentary Generator shares the memory-to-language projector
\(P_\theta\), language model, and decoding process across the three tracks,
while each track uses a track-specific context configuration. Let
\(\mathcal B_t^{\mathrm{sel}}\subseteq\mathcal B_t\) denote the selected local
completed-event memories at time \(t\), and let \([\cdot\,;\cdot]\) denote
memory concatenation. The latent soft prefix is
\begin{equation}
Z_t=
\begin{cases}
P_\theta(M_j^{\mathrm c}), &
\text{current-event},\\
P_\theta\!\left([M_t;\mathcal B_t^{\mathrm{sel}}]\right), &
\text{recent/historical}.
\end{cases}
\label{eq:commentary_context}
\end{equation}
Current-event commentary uses the just-closed completed-event memory
\(M_j^{\mathrm c}\). Recent-window and historical-memory commentary concatenate
the active event memory \(M_t\) with selected local completed-event memories and
compress them with \(P_\theta\) into a fixed-length soft prefix.
Historical-memory commentary additionally serializes retrieved records
\(\mathcal R_t\subseteq\mathcal L_t\) as prompt text, allowing the shared
generator to combine current and recent latent context with retrieved historical
record text.

\subsection{Online Runtime and Rule-Assisted Scheduling}

At decision step \(t\), long-term retrieval filters \(\mathcal L_t\) by temporal
range and ranks eligible records with the LTM retrieval query projection,
returning the top-\(k\) records \(\mathcal R_t\). Given the just-completed event,
local memories, retrieved records, and commentary cadence, a deterministic
rule-assisted scheduler selects
\(\delta_t\in\{\text{silence},\text{current-event},\allowbreak
\text{recent-window},\text{historical-memory}\}\).

Current-event mode requires a salient just-completed event and a satisfied
cooldown; recent-window mode requires sufficient completed events and passes
periodic and event-proximity gates; historical-memory mode requires records
that satisfy temporal, relevance, and cooldown conditions. Otherwise, the
system remains silent. When multiple modes are eligible, a fixed priority
selects one mode and its context configuration; non-silent decisions invoke
the Multi-Context Commentary Generator. The learned query projection only
ranks historical candidates, whereas the rule-assisted policy determines
speaking time, commentary mode, and context configuration. The scheduler uses
fixed thresholds, priorities, cooldowns, and queue parameters, while
evaluation-specific interventions are defined by the experimental protocol.

\begin{table*}[!t]
\centering
\footnotesize
\setlength{\tabcolsep}{2.0pt}
\renewcommand{\arraystretch}{1.06}
\begin{tabular*}{\textwidth}{@{\extracolsep{\fill}}cccc|ccc|ccc|ccc@{}}
\toprule
\multirow{2}{*}{Method} &
\multirow{2}{*}{Frames} &
\multirow{2}{*}{Stream} &
\multirow{2}{*}{Soccer FT} &
\multicolumn{3}{c|}{Current-event} &
\multicolumn{3}{c|}{Recent-window} &
\multicolumn{3}{c}{Historical-memory} \\
\cmidrule(lr){5-7}\cmidrule(lr){8-10}\cmidrule(lr){11-13}
& & & & CIDEr & B@4 & BS & CIDEr & B@4 & BS & CIDEr & B@4 & BS \\
\midrule
\multicolumn{13}{c}{\bfseries Proprietary MLLMs} \\
\cmidrule(lr){1-13}
GPT-5.4 &
16 &
&
&
5.93 & 0.0215 & 0.8586 &
14.28 & 0.0585 & \underline{0.8784} &
\underline{7.75} & \underline{0.0405} & \underline{0.8723} \\
Qwen3.6-Max-Preview &
16 &
&
&
1.00 & 0.0092 & \underline{0.8613} &
3.23 & 0.0131 & 0.8576 &
3.95 & 0.0194 & 0.8655 \\
\midrule
\multicolumn{13}{c}{\bfseries Streaming VLMs} \\
\cmidrule(lr){1-13}
StreamingVLM (ICLR2026) &
2 FPS &
 $\checkmark$ &
&
0.7351 & 0.0018 & 0.4337 &
0.0532 & 0.0015 & 0.4398 &
0.0313 & 0.0000 & 0.4464 \\
TimeChat-Online (ACM MM2025) &
2 FPS &
 $\checkmark$ &
&
1.2771 & 0.0000 & 0.4042 &
0.9740 & 0.0000 & 0.4685 &
1.0078 & 0.0000 & 0.4494 \\
VideoLLM-online (CVPR2024) &
2 FPS &
 $\checkmark$ &
&
0.1595 & 0.0000 & 0.2345 &
0.0443 & 0.0000 & 0.2008 &
0.0477 & 0.0000 & 0.2393 \\
\midrule
\multicolumn{13}{c}{\bfseries Soccer-Specific Models} \\
\cmidrule(lr){1-13}
UniSoccer (CVPR2025) &
30 &
&
 $\checkmark$ &
30.97 & \underline{0.2719} & 0.6782 &
\textbf{27.85} & \textbf{0.0730} & 0.6510 &
5.69 & 0.0317 & 0.5679 \\
SoccerMaster (CVPR2026) &
30 &
&
 $\checkmark$ &
\underline{33.35} & 0.2650 & 0.6827 &
23.46 & 0.0569 & 0.5984 &
7.37 & 0.0273 & 0.5418 \\
StreamSoccer (Ours) &
2 FPS &
 $\checkmark$ &
 $\checkmark$ &
\textbf{38.62} & \textbf{0.2929} & \textbf{0.9734} &
\underline{23.96} & \underline{0.0652} & \textbf{0.9612} &
\textbf{17.39} & \textbf{0.0451} & \textbf{0.9636} \\
\bottomrule
\end{tabular*}
\caption{Commentary quality at shared causal output anchors. We report CIDEr,
BLEU-4 (B@4), and BERTScore-F1 (BS) for current-event, recent-window, and
historical-memory commentary; higher is better. CIDEr uses the \(100\times\)
scale. Bold and underlined values denote the best and second-best result for
each metric within each track.}
\label{tab:common_anchor_quality}
\end{table*}

\subsection{Training Objectives}

StreamSoccer uses three training stages for event-memory learning,
memory-to-record consolidation, and multi-context commentary generation. The
visual encoder and pretrained language-model backbone remain frozen, with
language-side adaptation implemented through LoRA \citep{hu2022lora}.

Stage~1 trains the Streaming Event Memory Encoder with four complementary
objectives:
\begin{equation}
\begin{aligned}
\mathcal L_{\mathrm{Stage1}}={}&
\alpha_{\mathrm{clip}}\mathcal L_{\mathrm{clip}}
+\alpha_{\mathrm{tr}}\mathcal L_{\mathrm{tr}}\\
&+\alpha_{\mathrm{evt\text{-}act}}\mathcal L_{\mathrm{evt\text{-}act}}
+\alpha_{\mathrm{evt\text{-}type}}\mathcal L_{\mathrm{evt\text{-}type}}.
\end{aligned}
\label{eq:stage1_objective}
\end{equation}
Here, \(\mathcal L_{\mathrm{clip}}\) uses asymmetric loss
\citep{ridnik2021asymmetric} for sparse multi-label action recognition on every
clip, and \(\mathcal L_{\mathrm{tr}}\) uses focal binary cross-entropy
\citep{lin2017focal} for operational-transition prediction at valid positions.
Upon event completion, \(\mathcal L_{\mathrm{evt\text{-}act}}\) uses focal
binary cross-entropy to supervise the multi-label action set, while
\(\mathcal L_{\mathrm{evt\text{-}type}}\) uses cross-entropy for the coarse
event type. Duration-safeguard positions are masked from operational-transition
supervision.

Stage~2 trains the memory-to-language projector and language-model LoRA
parameters to generate \(c_j\) from \(M_j^{\mathrm c}\). Its autoregressive
objective \(\mathcal L_{\mathrm{Stage2}}\) is applied only to assistant caption
tokens.

Stage~3 is initialized from the first two stages and optimizes the trainable
modules enabled in the reported configuration over all three commentary
tracks. The task-balanced \(\mathcal L_{\mathrm{gen}}\) supervises commentary
generation, while \(\mathcal L_{\mathrm{retain}}\) reuses the four Stage~1
objectives to preserve event semantics. The memory-context objective
\(\mathcal L_{\mathrm{route}}\) supervises the required memory context,
\(\mathcal L_{\mathrm{local}}\) ranks recent candidate events, and
\(\mathcal L_{\mathrm{retr}}\) ranks historical candidate records. The latter
two ranking losses are computed only when valid positive and negative evidence
is available.

The complete Stage~3 objective is
\begin{equation}
\begin{aligned}
\mathcal L_{\mathrm{Stage3}}={}&
\lambda_{\mathrm{gen}}\mathcal L_{\mathrm{gen}}
+\lambda_{\mathrm{retain}}\mathcal L_{\mathrm{retain}}\\
&+\lambda_{\mathrm{route}}\mathcal L_{\mathrm{route}}
+\lambda_{\mathrm{local}}\mathcal L_{\mathrm{local}}
+\lambda_{\mathrm{retr}}\mathcal L_{\mathrm{retr}}.
\end{aligned}
\label{eq:stage3_objective}
\end{equation}
The three context-selection objectives provide auxiliary supervision for
memory use. At runtime, the rule-assisted scheduler determines the commentary
mode and local context selection, while the query projection learned through
\(\mathcal L_{\mathrm{retr}}\) ranks event records. This separates learned
context ranking from deterministic runtime decisions.

\section{Experiments}
\label{sec:experiments}

We evaluate StreamSoccer through common-anchor commentary quality, raw-video
efficiency as match history grows, event-memory representation and context
ablations, and a qualitative case study.

\subsection{Experimental Setup}

\paragraph{Datasets and Splits.}
We construct three-track streaming soccer commentary data from SoccerNet action
annotations and MatchTime commentary. The dataset contains 27,639 samples:
15,189 current-event, 7,127 recent-window, and 5,323 historical-memory examples.
Each sample has an explicit observation cutoff. The training, validation, and
test splits contain 19,641, 4,209, and 3,789 samples, respectively, and are
match-disjoint.

SoccerNet provides human-annotated action timestamps but not event intervals. We
use deterministic football rules based on action categories and temporal order
to organize consecutive clips into rule-derived operational events.
Action-triggered event starts form positive operational-transition targets,
while within-event clips form negatives. To bound the duration of a state, an
active event is closed after 24~s and the next clip initializes a new state.
This duration safeguard limits state length and is not treated as a positive
operational transition.

Event-focused MatchTime descriptions are aligned to operational events by
match, half, temporal proximity, and action compatibility, with each description
assigned to at most one event. The aligned text and deterministic time, action,
and event-type metadata form data-side reference records for three commentary
tracks: current-event targets rewrite an aligned description as post-event
commentary; recent-window targets use only video and completed local events
available before the observation cutoff; and historical-memory targets
additionally associate earlier eligible events from the same match and half.
Data-side reference records are used only to construct supervision, whereas
runtime event records are generated from completed event memory. Automated
checks ensure that all conditioning information is available before the
observation cutoff and retain the corresponding source-event identifiers.

\paragraph{Evaluation Design.}
Common-anchor evaluation fixes the task type, observation cutoff, and reference
emission time, and discloses current-event alignment and record-readiness
interventions. Raw-video evaluation additionally includes video decoding,
online visual encoding, and runtime prediction. Mechanism experiments
distinguish retrained structural variants from inference-time interventions.

\paragraph{Baselines and Metrics.}
Strict comparisons share the test split, cutoff, output anchor, input scope, and
decoding budget; results that cannot satisfy this contract are reported as
compatibility comparisons. We report track-wise BLEU-4, CIDEr
\citep{vedantam2015cider}, and BERTScore-F1 \citep{zhang2020bertscore}. We
compute BERTScore-F1 for all methods using RoBERTa-large at layer 17.
Memory-to-record representation experiments additionally report token-level
F1, METEOR, and ROUGE-L. Raw-video efficiency is measured by the
95th-percentile real-time factor (RTF) of the per-minute workload.

\FloatBarrier
\subsection{Commentary Quality}

Table~\ref{tab:common_anchor_quality} compares inference at common reference
anchors.

The proprietary MLLMs receive 16 track-specific sparse RGB frames per
anchor. StreamingVLM, TimeChat-Online, and VideoLLM-online process the timelines
at 2 FPS, with metrics conditioned on valid outputs. UniSoccer and SoccerMaster
are retrained on match-disjoint current-event and recent-window data and receive
30 causal frames; their historical rows are test-only sparse-history
diagnostics without historical-memory training. StreamSoccer uses
reference-anchored replay over 126 half streams, with oracle operational-event
closures, current-event alignment, label-assisted action/type metadata, and
record-readiness interventions.

StreamSoccer obtains the highest reported CIDEr on current-event and
historical-memory commentary and ranks second on recent-window commentary,
behind UniSoccer. B@4 shows the same ordering on current-event and recent-window
commentary, while StreamSoccer also ranks first on the historical-memory track.
Recent multi-event aggregation remains its main relative weakness. The
historical rows of the soccer-specific baselines use untrained sparse-frame
histories and therefore serve as compatibility comparisons.

\subsection{Streaming Efficiency and Long-History Scaling}

We further examine whether the computational cost of the complete raw-video
path grows with the observed match history. Experiments measure the
95th-percentile real-time factor (RTF) of the per-minute workload over 174 runs
on 58 source-clean matches. All methods receive the same 2 FPS input; an RTF
below 1 indicates faster-than-real-time processing.

\begin{figure}[!t]
\centering
\includegraphics[width=\columnwidth]{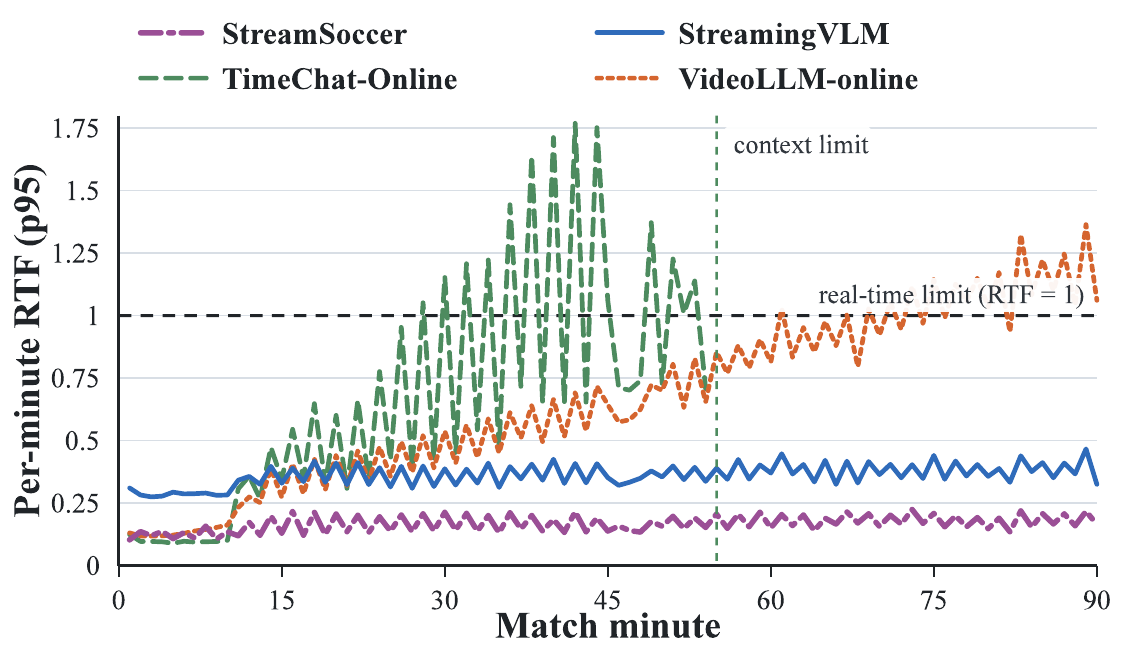}
\caption{Raw per-minute RTF p95 over 58 source-clean matches and 174 runs
(lower is better). TimeChat-Online stops at its context limit.}
\label{fig:raw_video_rtf}
\end{figure}

As shown in Figure~\ref{fig:raw_video_rtf}, StreamSoccer maintains an RTF p95 of
approximately 0.10--0.22 throughout a 90-minute match, without sustained growth
as more match history is observed. StreamingVLM remains at approximately
0.31--0.47. VideoLLM-online becomes increasingly expensive as its context grows,
approaching or exceeding the real-time threshold late in the match, whereas
TimeChat-Online reaches its context limit at approximately 55 minutes. These
results show that StreamSoccer maintains stable, faster-than-real-time
processing as the observed match history grows.

\FloatBarrier

\subsection{Validating Event Memory as an Intermediate Representation}

To evaluate event memory as an intermediate representation for event-record
caption generation, we compare recurrent event memory with operational-event
pooling and fixed-time pooling under matched visual inputs and language
readout. Operational-event pooling shares the same event support as recurrent
memory, providing a matched comparison of recurrent state formation, while
fixed-time pooling serves as a temporal-window baseline.

\begin{table}[!t]
\centering
\small
\setlength{\tabcolsep}{2.0pt}
\begin{tabular}{@{}lccccc@{}}
\toprule
Representation & Tok-F1 & B@4 & METEOR & ROUGE-L & CIDEr \\
\midrule
Recurrent memory
    & \textbf{0.2642} & \textbf{0.2777} & \textbf{0.2358}
    & \textbf{0.4335} & \textbf{37.40} \\
Event pooling
    & 0.2290 & 0.2556 & 0.2192 & 0.3986 & 22.20 \\
Fixed-time pooling
    & 0.2108 & 0.2462 & 0.2104 & 0.3916 & 15.72 \\
\bottomrule
\end{tabular}
\caption{Matched projector-only comparison for memory-to-record consolidation.}
\label{tab:event_representation}
\end{table}

Table~\ref{tab:event_representation} shows a consistent ordering across all
caption metrics. Recurrent event memory improves Tok-F1 by 0.0533 over
fixed-time pooling and by 0.0351 over operational-event pooling, with CIDEr
gains of 21.68 and 15.21, respectively. Operational-event pooling also improves
Tok-F1 by 0.0182 over fixed-time pooling, showing the benefit of organizing
visual support around operational events rather than a fixed trailing window.
With the event support held fixed, recurrent event memory provides a further
consistent gain, indicating that recurrent integration of event evolution
forms a more useful intermediate representation for downstream generation.

\begin{figure}[!ht]
\centering
\includegraphics[width=0.98\columnwidth]{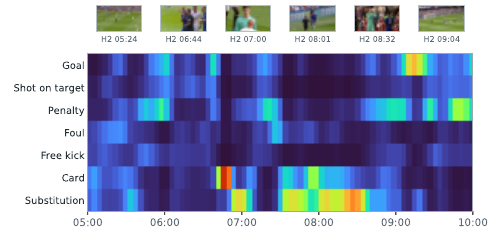}
\caption{Normalized commentary-conditioned temporal relevance from a frozen
offline VLM over a continuous five-minute match interval. Responses persist
across adjacent windows and shift among action concepts as play evolves.}
\label{fig:event_relevance}
\end{figure}
\FloatBarrier

\paragraph{Memory-Scope Analysis.}

We compare task adaptation and memory scope in two blocks. Direct Qwen3-VL
reads causal 30/180/180\,s video windows;
Qwen3-VL Video-SFT retains this interface and is adapted on the three-track
training set. Three separately trained StreamSoccer variants then use current
event memory only, add the local event-memory buffer, or further add historical event
records. The direct-video and event-memory blocks assess task adaptation and
memory scope, respectively.

\begin{table}[!ht]
\centering
\footnotesize
\setlength{\tabcolsep}{2.4pt}
\begin{tabular}{@{}lccc@{}}
\toprule
Variant & Current & Recent & Historical \\
\midrule
Direct Qwen3-VL & 4.06 & 2.46 & 3.08 \\
Qwen3-VL Video-SFT & 27.73 & 12.98 & 11.24 \\
\midrule
\multicolumn{4}{c}{\itshape StreamSoccer variants} \\
Current memory only & 33.03 & 15.06 & 13.22 \\
\(\quad+\) Local event-memory buffer
    & \underline{38.57} & \underline{17.96} & \underline{17.22} \\
\(\quad+\) Historical records (Full)
    & \textbf{38.62} & \textbf{23.96} & \textbf{17.39} \\
\bottomrule
\end{tabular}
\caption{Track-wise CIDEr for direct-video adaptation and event-memory scope.}
\label{tab:memory_scope_analysis}
\end{table}
\FloatBarrier

Video-SFT raises CIDEr from 4.06/2.46/3.08 to 27.73/12.98/11.24 across the
three tracks, showing the importance of task adaptation. Within the
event-memory block, adding the local event-memory buffer improves all three tracks
over current memory alone. The full configuration further raises the scores to
38.62/23.96/17.39 and performs best on every track, with the largest additional
gain on recent-window commentary. Together, the three contexts support
commentary over different temporal ranges.

\begin{figure}[!ht]
\centering
\includegraphics[width=\columnwidth]{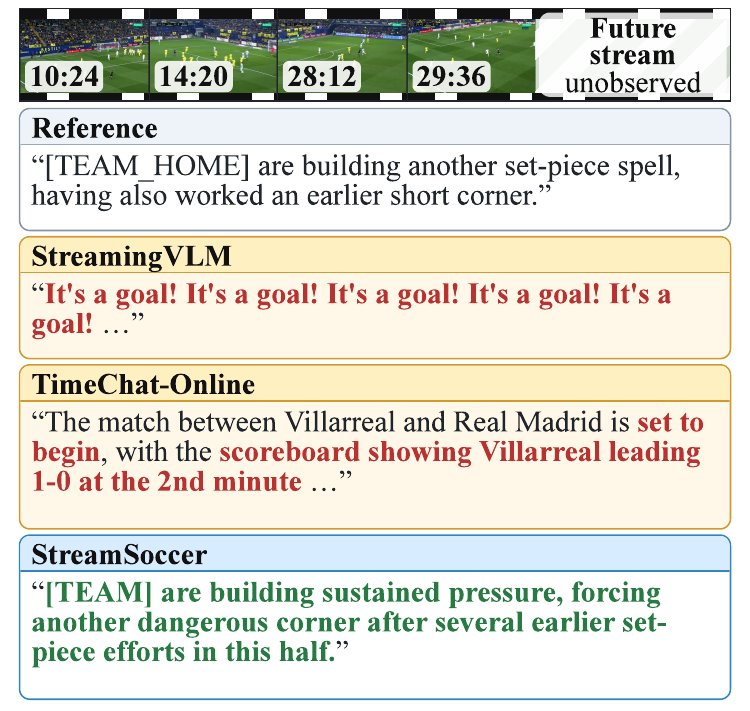}
\caption{Historical-memory commentary at a common causal anchor. The masked
future stream is unobserved.}
\label{fig:historical_case}
\end{figure}
\FloatBarrier

\subsection{Qualitative Case Analysis}

Figure~\ref{fig:historical_case} compares historical-memory commentary at a
common causal anchor containing both earlier set-piece context and an ongoing
attacking sequence.

StreamingVLM produces a repeated goal claim and fails to organize the current
play, while TimeChat-Online generates descriptions inconsistent with the match
time and score. StreamSoccer instead connects the ongoing pressure with earlier
set-piece events and more closely matches the reference. By retaining current,
recent, and historical event contexts beyond a finite video window, the layered
memory reduces abrupt context loss as the window advances and supports more
coherent commentary.

\FloatBarrier
\section{Conclusion}

We introduced StreamSoccer, an event-driven system for causal streaming soccer
commentary. It maintains fixed-budget active event memory, preserves recent
completed events, and consolidates them into retrievable event records to
support current-event, recent-window, and historical-memory commentary. At
common output anchors, StreamSoccer achieves the highest reported CIDEr on the
current-event and historical-memory tracks, while its raw-video RTF remains
below the real-time threshold throughout full matches. Memory-scope analysis
further shows that recent completed events improve all three commentary tracks
and that the full configuration performs best overall. Together with the
representation comparison and long-history efficiency results, these findings
support event memory as a compact intermediate representation for long
streaming soccer video.

\FloatBarrier
\bibliographystyle{aaai2027}

{\small
\bibliography{aaai2027}
}

\clearpage
\appendix
\makeatletter
\@addtoreset{figure}{section}
\@addtoreset{table}{section}
\@addtoreset{algorithm}{section}
\setlength{\@fptop}{0pt}
\setlength{\@dblfptop}{0pt}
\makeatother
\setcounter{figure}{0}
\setcounter{table}{0}
\setcounter{algorithm}{0}
\renewcommand{\thefigure}{\thesection.\arabic{figure}}
\renewcommand{\thetable}{\thesection.\arabic{table}}
\renewcommand{\thealgorithm}{\thesection.\arabic{algorithm}}

\section{Evidence and Evaluation Contract}
\label{app:evidence_contract}

The paper uses separate protocols for controlled commentary quality, streaming
state execution, and raw-video efficiency. Keeping these protocols separate
prevents a controlled output-anchor experiment from being interpreted as an
evaluation of autonomous speaking decisions.

\subsection{Claim--Evidence Map}

Table~\ref{tab:claim_evidence} records what each headline experiment supports
and, equally importantly, what it does not test.

\begin{table*}[t]
\centering
\small
\setlength{\tabcolsep}{5pt}
\renewcommand{\arraystretch}{1.08}
\begin{tabular}{@{}p{0.20\textwidth}p{0.24\textwidth}p{0.24\textwidth}p{0.23\textwidth}@{}}
\toprule
Paper claim & Primary evidence & Supported interpretation & Outside the experiment's scope \\
\midrule
Commentary quality at shared causal anchors
& Reference-anchored streaming replay and common-anchor baselines
& Conditional generation quality at fixed legal output anchors for the three commentary tracks
& Free speaking-time selection, naturally available event records, or a purely visual front end \\
Availability-aware semantic preference at shared anchors
& Frozen four-way blinded LLM audit against silver references
& Reference-conditioned judge preference over the fixed 3,789-anchor population, including missing-output penalties
& Human preference, claim-level factuality, historical-record attribution, free speaking-time selection, or judge-model generalization \\
Event memory is a useful intermediate representation
& Matched recurrent-memory, operational-event-pooling, and fixed-time-pooling comparison
& Downstream event-record caption decodability under a matched language readout and adaptation budget
& Natural event discovery, full record factuality, or equal upstream training cost \\
Accessible memory scope changes track-specific commentary quality
& Separately trained memory-scope variants in the main text
& Observed differences between configurations with different accessible memory scopes and training targets
& Statistical significance of every incremental gain or a causal decomposition of every component \\
Long-history processing remains efficient
& Raw-video per-minute RTF scaling over full matches
& Growth of end-to-end processing workload as more match history is observed
& Scheduler quality or the response latency of an individual commentary utterance \\
\bottomrule
\end{tabular}
\caption{Claim--evidence map for the headline results. Each row deliberately
limits the interpretation to the corresponding evaluation protocol.}
\label{tab:claim_evidence}
\end{table*}

\subsection{Evaluation Protocol Matrix}

We distinguish four evaluation settings in Table~\ref{tab:protocol_matrix}.
\emph{Sample-level conditional evaluation} reads a preconstructed causal
sample and is used for model-development diagnostics. \emph{Reference-anchored
streaming replay} advances persistent state over each half but takes the output
anchor and task from the test row. \emph{Free-scheduler streaming replay}
advances the same type of state while allowing the deterministic rule-assisted
scheduler to choose a commentary mode or silence. \emph{Raw-video scaling}
includes video decoding and online visual encoding and measures system cost as
the visible history grows.

\begin{table*}[t]
\centering
\scriptsize
\setlength{\tabcolsep}{2.5pt}
\renewcommand{\arraystretch}{1.10}
\begin{tabular}{@{}p{0.105\textwidth}p{0.13\textwidth}p{0.11\textwidth}p{0.11\textwidth}p{0.12\textwidth}p{0.105\textwidth}p{0.12\textwidth}p{0.105\textwidth}@{}}
\toprule
Setting & Visual input and state & Output time / task & Event closure & Action/type metadata & Current-event alignment & Record readiness & Main use \\
\midrule
Sample-level conditional
& Preconstructed or cached sample; no full-half replay required
& Given by sample
& Controlled or materialized
& Controlled fields
& Given by sample IDs
& Preconstructed evidence context
& Conditional generation and matched ablations \\
Reference-anchored replay
& Continuous cached Qwen tokens at 2 FPS; state persists within each half
& Reference anchor and task type
& Oracle operational closures
& Label-assisted action/type fields
& Just-completed event aligned to the current-event target
& Historical anchors deterministically drain pending caption jobs before retrieval
& Main-paper common-anchor commentary quality \\
Free-scheduler replay
& Continuous cached Qwen tokens; state persists within each half
& Rule-assisted scheduler
& Predicted transition plus duration rollover
& Current audited replay is label-assisted
& None from reference targets
& Natural asynchronous readiness
& Cadence, mode counts, delay, duplicates, and readiness audit \\
Raw-video scaling
& Raw video decoding, online visual encoding, persistent state
& Fixed evaluation workload; separate from native free scheduling
& Predicted transition plus duration rollover
& Predicted front-end metadata
& None from reference targets
& Runtime-managed records
& Per-minute RTF, VRAM, coverage, and history scaling \\
\bottomrule
\end{tabular}
\caption{Evaluation protocol matrix. Table~\ref{tab:common_anchor_quality} uses
the reference-anchored row. It evaluates commentary quality under controlled
anchors; it does not evaluate the speaking scheduler or natural record
readiness.}
\label{tab:protocol_matrix}
\end{table*}

The frozen reference-anchored run behind the StreamSoccer row of
Table~\ref{tab:common_anchor_quality} uses checkpoint step 1,000. It processes
86,145 causal clips from 63
test matches (126 half streams), emits all 3,789 test anchors, and constructs
20,689 event-record traces. Inference uses cached Qwen tokens, oracle
operational closures, rule memory gating, embedding-based top-3 retrieval,
same-half eligibility, and a 90-s minimum historical gap. Current-event
identity alignment and record-readiness intervention are evaluation controls.
The result establishes controlled streaming quality, not free-scheduler system
quality.

\subsection{Baseline and Statistical Contract}

The external rows in the main table have different native interfaces. Table
\ref{tab:baseline_cards} makes the consequential differences explicit. We call
a comparison \emph{strict} only when the test split, causal cutoff, output
anchor, input scope, and decoding budget can be aligned. A row that preserves
the method's native interface but cannot meet all of these conditions is a
\emph{compatibility comparison}; it is retained for context and does not
constitute a controlled architectural ranking.

\begin{table*}[t]
\centering
\scriptsize
\setlength{\tabcolsep}{3.0pt}
\renewcommand{\arraystretch}{1.08}
\begin{tabular}{@{}p{0.17\textwidth}cccp{0.18\textwidth}p{0.20\textwidth}p{0.13\textwidth}@{}}
\toprule
Method family & Input & Stream state & Soccer FT & Anchor/interface & Coverage treatment & Contract \\
\midrule
Proprietary MLLMs
& 16 frames & No & No
& Track-specific causal sparse frames at each anchor
& One request per test anchor; invalid calls must be reported
& Compatibility \\
General streaming VLMs
& 2 FPS & Yes & No
& Continuous causal timeline with anchored scoring
& Metrics are conditioned on valid outputs; context-limit and invalid-output coverage are disclosed separately
& Compatibility \\
Soccer-specific models
& 30 frames & No & Yes
& Causal sparse frames at each anchor
& Retrained current/recent rows use the match-disjoint split; historical rows are test-only sparse-history diagnostics
& Current/recent stricter; historical compatibility \\
StreamSoccer
& 2 FPS & Yes & Yes
& Reference-anchored continuous replay
& 3,789/3,789 anchors emitted; interventions disclosed in Table~\ref{tab:protocol_matrix}
& Controlled reference replay \\
\bottomrule
\end{tabular}
\caption{Protocol cards for the method families in the main commentary-quality
table. FT denotes task-specific fine-tuning.}
\label{tab:baseline_cards}
\end{table*}

CIDEr is reported on the \(100\times\) scale. BERTScore-F1 uses the shared
RoBERTa-large scorer at layer 17, and all generated-output rows are rescored by
the same evaluation code. Uncertainty is clustered by match because anchors
from the same match are not independent. For a paired comparison we resample
matches with replacement and retain all anchors belonging to the sampled
matches. When only one training seed is available, the resulting confidence
interval quantifies test-set variation, not training variance; we report the
checkpoint and seed rather than implying multi-seed stability. The frozen
representation comparison in Appendix~\ref{app:event_memory_evidence}
uses 10,000 paired match-cluster bootstrap replicates.

\subsection{Metrics and Interpretation Boundaries}

Table~\ref{tab:metric_definitions} groups the reported metrics by the question
they answer. Text-reference similarity, retrieval, continuous execution, and
raw-video efficiency are evaluated separately; a strong value in one group
does not substitute for evidence in another.

\begin{table*}[t]
\centering
\scriptsize
\setlength{\tabcolsep}{4pt}
\renewcommand{\arraystretch}{1.08}
\begin{tabular}{@{}p{0.15\textwidth}p{0.34\textwidth}p{0.42\textwidth}@{}}
\toprule
Metric & Definition & Interpretation boundary \\
\midrule
Token F1
& Harmonic mean of token-level precision and recall against the reference
& Direct lexical-content overlap; insensitive to unsupported facts expressed with overlapping words \\
BLEU-4
& Modified \(1\)- to \(4\)-gram precision with a brevity penalty
& Local phrase agreement; brittle to valid paraphrases and not a factuality metric \\
METEOR
& Unigram alignment combining precision, recall, and an ordering penalty
& More recall-sensitive than BLEU, but still reference-dependent \\
ROUGE-L
& Longest-common-subsequence overlap between prediction and reference
& Sequence-level content coverage and ordering, not historical-evidence use \\
CIDEr
& TF--IDF-weighted \(n\)-gram agreement; displayed on the \(100\times\) scale
& Main caption-style content score; scales and scorer versions must not be mixed \\
BERTScore-F1
& Contextual-token similarity under the shared RoBERTa-large layer-17 scorer
& Semantic reference similarity; does not establish factual correctness \\
Valid coverage
& Non-empty valid outputs divided by requested output anchors
& Must accompany quality scores so failed requests are not silently removed \\
Strict blind win rate
& Fraction of the fixed test population for which an anonymized candidate is selected under the fixed judge configuration; \texttt{NO\_RESPONSE} cannot win
& Availability-aware preference against one silver reference under one judge, not human-verified semantic correctness \\
R@\(k\) / MRR
& Fraction of eligible positives in the top \(k\), and mean reciprocal positive rank
& Retrieval ranking on the declared eligible population; does not show that generation used the record \\
Retrieval-ready rate
& Decisions at which at least one eligible, fully inserted event record is available
& Runtime record availability; distinct from retrieval correctness \\
RTF
& Processing wall time divided by represented video duration
& Values below one are faster than real time; cached-replay and raw-video RTFs are not interchangeable \\
Latency / VRAM
& Request p50/p95 wall time and peak process memory
& Tail response and resource cost under the stated timing boundary and valid-run population \\
Clustered 95\% CI
& Bootstrap interval obtained by resampling matches and retaining their anchors
& Test-population variation; with one training seed it is not training-seed uncertainty \\
\bottomrule
\end{tabular}
\caption{Metric definitions and permitted interpretations. No automatic
reference-similarity metric is treated as a direct factuality measure.}
\label{tab:metric_definitions}
\end{table*}

\FloatBarrier
\section{Data Construction and Operational-Event Semantics}
\label{app:data_events}

\subsection{Sources, Funnel, and Match-Disjoint Splits}

We use SoccerNet full-match videos and action timestamps
\citep{giancola2018soccernet,deliege2021soccernetv2} to construct the causal
video stream and operational events. MatchTime commentary
\citep{rao2024matchtime} provides the text used for the three-track commentary
data. SoccerNet-Caption \citep{mkhallati2023soccernetcaption} supplies the
legacy memory-to-record representation benchmark used in
Appendix~\ref{app:event_memory_evidence}. The roles and frozen data funnel are
shown in Table~\ref{tab:data_funnel}.

\begin{table}[t]
\centering
\small
\setlength{\tabcolsep}{5pt}
\begin{tabular}{@{}lr@{}}
\toprule
Data level & Count \\
\midrule
SoccerNet matches / half streams & 500 / 1,000 \\
Non-overlapping 4-s clips & 685,903 \\
Rule-derived operational events & 166,419 \\
MatchTime matches used & 422 \\
Caption-aligned operational events & 15,189 \\
Current-event samples & 15,189 \\
Recent-window samples & 7,127 \\
Historical-memory samples & 5,323 \\
Total three-track samples & 27,639 \\
\bottomrule
\end{tabular}
\caption{Frozen data funnel. Commentary alignment covers a salient subset of
all operational events rather than a uniform sample of match states.}
\label{tab:data_funnel}
\end{table}

The 422 MatchTime matches are split at match level using the sorted match ID:
70\% train, 15\% validation, and 15\% test. Table~\ref{tab:data_splits} shows
the resulting counts. The canonical audit finds no match or sample-ID overlap
across splits. Every sample stores a causal cutoff and provenance event IDs;
automated checks reject video, local-event, or historical references that form
after that cutoff.

\begin{table}[t]
\centering
\small
\setlength{\tabcolsep}{3.2pt}
\begin{tabular}{@{}lrrrrr@{}}
\toprule
Split & Matches & Current & Recent & Historical & Total \\
\midrule
Train & 295 & 10,853 & 5,013 & 3,775 & 19,641 \\
Validation & 64 & 2,286 & 1,105 & 818 & 4,209 \\
Test & 63 & 2,050 & 1,009 & 730 & 3,789 \\
\midrule
Total & 422 & 15,189 & 7,127 & 5,323 & 27,639 \\
\bottomrule
\end{tabular}
\caption{Match-disjoint three-track split.}
\label{tab:data_splits}
\end{table}

\FloatBarrier
\subsection{Operational Eventization and Boundary Timeline}

Each half is partitioned into non-overlapping four-second clips. For data
assignment, clip \(C_t\) uses the half-open interval \([s_t,e_t)\), so an action
timestamp \(a\) belongs to \(C_t\) iff \(s_t\le a<e_t\). The causal model may
use \(C_t\) only after \(e_t\le\tau_t\). At 2 FPS, a full clip contains eight
sampled frames.

Operational eventization groups the SoccerNet action timestamps using
deterministic football rules. Set pieces and administrative actions can start
a new event. Finish, stoppage, administrative, and eligible clearance actions
can close the active event after the triggering clip has been included. A
maximum duration \(D_{\max}=24\) s closes an overly long state, and the last
active state is deterministically closed at half end. These are operational
units for memory control; they are not manually segmented natural events.

Table~\ref{tab:event_action_groups} gives the exact action groups used by the
eventizer. When a clip contains several actions, a fixed priority resolves the
primary action: goals and shots precede penalties and cards, followed by
fouls/offside, set pieces, clearance and ball-out actions, substitutions,
kick-off, and end-of-period control labels. A card that coincides with a
stoppage is retained in the stoppage episode rather than forcing a redundant
start.

\begin{table*}[t]
\centering
\small
\setlength{\tabcolsep}{4pt}
\renewcommand{\arraystretch}{1.08}
\begin{tabular}{@{}p{0.18\textwidth}p{0.43\textwidth}p{0.31\textwidth}@{}}
\toprule
Group & SoccerNet actions & Operational role \\
\midrule
Set piece
& Corner, Throw-in, Direct free-kick, Indirect free-kick, Penalty, Kick-off
& Closes an existing state before \(C_t\) and starts a new event at \(C_t\) \\
Finish
& Goal, Shots on target, Shots off target
& Included in the active event, then closes it as \texttt{terminal\_finish} \\
Stoppage
& Ball out of play, Foul, Offside
& Included in the active event, then closes it as \texttt{terminal\_stoppage} \\
Defensive resolution
& Clearance
& Closes after \(C_t\) only when the next clip contains no set piece, finish, or stoppage follow-up \\
Administration / discipline
& Substitution, Yellow card, Red card, Yellow-to-red card, End of period
& Starts a new event unless a card coincides with an active stoppage; closes after inclusion \\
Open play
& No action from the groups above
& Continues the active state; its type is upgraded when the first meaningful action arrives \\
\bottomrule
\end{tabular}
\caption{Deterministic action groups and their operational-event effects.
These rules organize memory lifecycles; they do not define natural event
ground truth.}
\label{tab:event_action_groups}
\end{table*}

\begin{algorithm*}[t]
\caption{Rule-derived operational eventization for one half}
\label{alg:eventization}
\begin{algorithmic}[1]
\Require temporally ordered clips \(C_1,\ldots,C_T\), assigned action timestamps,
maximum duration \(D_{\max}\)
\Ensure ordered operational-event episodes
\State \(\mathit{active} \gets \varnothing\)
\For{each clip \(C_t\) in temporal order}
    \State \(\mathit{actions} \gets \Call{AssignedActions}{C_t}\)
    \State \(\mathit{next\_actions} \gets \Call{AssignedActions}{C_{t+1}}\) if \(t<T\), else \(\varnothing\)
    \If{\(\mathit{active}=\varnothing\)}
        \State \(\mathit{active} \gets \Call{StartEvent}{C_t}\)
    \ElsIf{$\Call{RequestsNewEvent}{C_t,\mathit{actions}}$}
        \State \Call{CloseEvent}{active}
        \State \(\mathit{active} \gets \Call{StartEvent}{C_t}\)
    \EndIf
    \State \(\Call{Attach}{\mathit{active}, C_t, \mathit{actions}}\)
    \State \Call{UpdateEventType}{active, actions}
    \If{$\Call{HasTerminalAction}{C_t,\mathit{actions},\mathit{next\_actions}}$}
        \State \Call{CloseEvent}{active}
        \State \(\mathit{active} \gets \varnothing\)
    \ElsIf{$\Call{Duration}{\mathit{active}} \ge D_{\max}$}
        \State \Call{CloseEvent}{active, \texttt{max\_duration}}
        \State \(\mathit{active} \gets \varnothing\)
    \EndIf
\EndFor
\If{\(\mathit{active}\neq\varnothing\)}
    \State \Call{CloseEvent}{active, \texttt{half\_end}}
\EndIf
\end{algorithmic}
\end{algorithm*}

Here, \textsc{RequestsNewEvent} covers set pieces and administration/discipline
actions except a card coincident with an active stoppage, while
\textsc{HasTerminalAction} covers finish, stoppage,
administration/discipline, and clearance without an immediate set-piece,
finish, or stoppage follow-up. Annotated next-clip actions are used only to
construct offline episode labels and are not model inputs. A start-triggered clip belongs to the new event; a
terminal-action clip remains in the event that it closes.

\begin{figure*}[t]
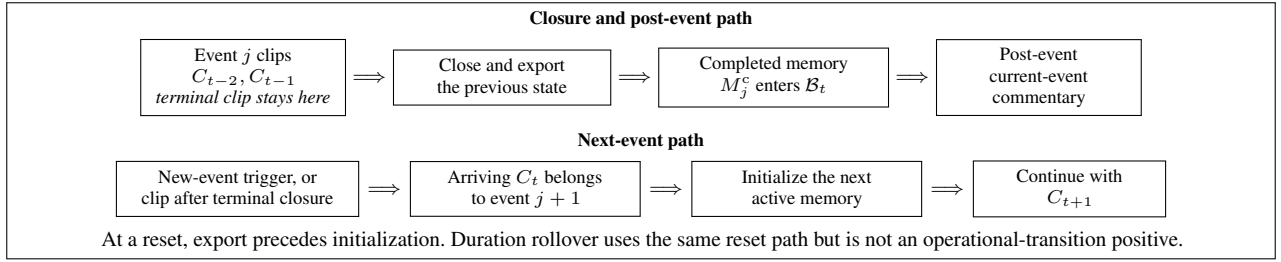

\centering
\fbox{\begin{minipage}{0.93\textwidth}
\centering
\scriptsize
\textbf{Closure and post-event path}

\vspace{0.45em}
\begin{tabular}{c@{\hspace{0.35em}}c@{\hspace{0.35em}}c@{\hspace{0.35em}}c@{\hspace{0.35em}}c@{\hspace{0.35em}}c@{\hspace{0.35em}}c}
\fbox{\parbox{0.15\textwidth}{\centering
Event \(j\) clips\\
\(C_{t-2}, C_{t-1}\)\\
\emph{terminal clip stays here}}}
& \(\Longrightarrow\) &
\fbox{\parbox{0.16\textwidth}{\centering
Close and export\\
the previous state}}
& \(\Longrightarrow\) &
\fbox{\parbox{0.17\textwidth}{\centering
Completed memory\\
\(M_j^{\mathrm c}\) enters \(\mathcal B_t\)}}
& \(\Longrightarrow\) &
\fbox{\parbox{0.15\textwidth}{\centering
Post-event\\
current-event commentary}}
\end{tabular}

\vspace{0.75em}
\textbf{Next-event path}

\vspace{0.45em}
\begin{tabular}{c@{\hspace{0.45em}}c@{\hspace{0.45em}}c@{\hspace{0.45em}}c@{\hspace{0.45em}}c@{\hspace{0.45em}}c@{\hspace{0.45em}}c}
\fbox{\parbox{0.18\textwidth}{\centering
New-event trigger, or\\
clip after terminal closure}}
& \(\Longrightarrow\) &
\fbox{\parbox{0.17\textwidth}{\centering
Arriving \(C_t\) belongs\\
to event \(j+1\)}}
& \(\Longrightarrow\) &
\fbox{\parbox{0.17\textwidth}{\centering
Initialize the next\\
active memory}}
& \(\Longrightarrow\) &
\fbox{\parbox{0.14\textwidth}{\centering
Continue with\\
\(C_{t+1}\)}}
\end{tabular}

\vspace{0.6em}
\footnotesize At a reset, export precedes initialization. Duration rollover
uses the same reset path but is not an operational-transition positive.
\end{minipage}}
\caption{Boundary and clip-ownership convention. The data eventizer stores the
transition label on the first clip of the new event. At reset, the pre-reset
state is the completed memory and the current clip initializes the next state.}
\label{fig:boundary_timeline}
\end{figure*}

The implementation places the transition label on the first clip of the new
event. Figure~\ref{fig:boundary_timeline} shows the ownership convention as a
flow. At a reset, the previous state is exported before the arriving clip
initializes the next event. A terminal-action clip remains in the event it
closes, and current-event supervision observes only the completed state.

The transition head predicts \(b_t\), the operational-transition indicator.
Training supplies a binary operational-transition label and masks
duration-rollover positions from that loss. \(\rho_t\) is the deterministic duration safeguard, and the final reset
is \(r_t=b_t\lor\rho_t\) (with deterministic stream-end closure handled
separately). Thus transition, rollover, and final reset are not interchangeable
labels.

The frozen eventizer produces 166,419 operational events with mean duration
16.49 s and median duration 20 s. A total of 66,941 events (40.2\%) close by
the maximum-duration safeguard. The large rollover fraction is why our claims
are consistently phrased in terms of \emph{rule-derived operational events},
not natural semantic-boundary annotations.

\subsection{Three-Track Supervision and Quality Control}

MatchTime commentary descriptions are aligned to operational events using match,
half, temporal proximity, and action compatibility, and each description is
assigned to at most one event. The aligned description and deterministic time,
action, event-type, and match-state fields form a data-side reference record.
This record is used to construct supervision; at runtime, historical records
are generated from completed event memory instead.

Candidate generation first restricts captions to the same match and half. A
caption is eligible when its timestamp lies inside the event interval or no
more than 15 s outside it. A second action-anchored path admits a caption up to
30 s from a compatible event action, accommodating delayed goal or shot
commentary while still requiring semantic action agreement. Candidates are
ranked deterministically by text validity, action compatibility, importance,
absence of multi-event ambiguity, availability and distance of an action
anchor, membership in or distance to the event window, annotation-label
availability, and finally caption ID. Semantically anchored high-confidence
candidates are assigned first across a match. In the frozen dataset, a caption
ID is consumed after assignment, so one caption supervises at most one canonical
event. Event IDs, action sets, timestamps, match state, and split membership
are fixed before any language rewriting.

\begin{table*}[t]
\centering
\footnotesize
\setlength{\tabcolsep}{6pt}
\renewcommand{\arraystretch}{1.14}
\begin{tabular}{@{}p{0.16\textwidth}p{0.76\textwidth}@{}}
\toprule
Track & Construction contract \\
\midrule
Current-event
& \textbf{Context:} just-completed \(M_j^{\mathrm c}\) and causal match state.
\textbf{Target:} one post-event commentary sentence.
\textbf{Provenance:} current event ID, cutoff, and alignment metadata. \\
Recent-window
& \textbf{Context:} active context and completed local events from the preceding
180 s. \textbf{Target:} one recent multi-event summary.
\textbf{Provenance:} selected local event IDs and cutoff. \\
Historical-memory
& \textbf{Context:} current/local context plus eligible earlier same-match,
same-half records. \textbf{Target:} one sentence relating current play to
earlier context. \textbf{Provenance:} historical source IDs, candidate set,
and cutoff. \\
\bottomrule
\end{tabular}
\caption{Construction contract for the three commentary tracks. Every
conditioning item must exist no later than the stored causal cutoff.}
\label{tab:track_construction}
\end{table*}

\begin{figure*}[b]
\centering
\includegraphics[width=\textwidth]{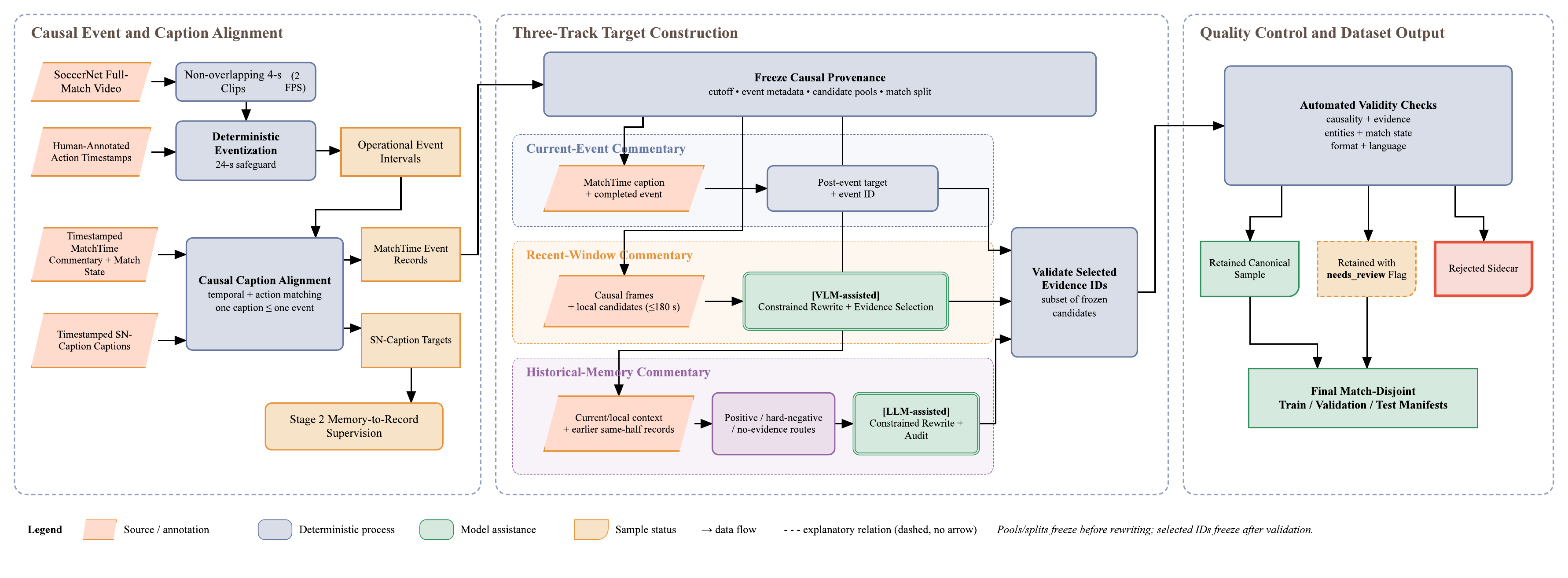}
\caption{Causal commentary-data construction. Action timestamps and
deterministic rules define operational-event intervals; MatchTime supplies the
main three-track branch and SoccerNet-Caption the Stage~2 supervision branch.
Candidate pools and split assignments are fixed before model-assisted
rewriting, while selected evidence IDs are validated before the final
match-disjoint manifests.}
\label{fig:data_construction_pipeline}
\end{figure*}

Automated quality control checks causal validity, match/half consistency,
event-ID existence, one-to-one current-caption assignment, duplicate sample
IDs, and split isolation. Text transformations preserve deterministic team,
time, action, and score fields outside unconstrained language generation. The
frozen automatic audit establishes structural consistency and traceability; it
does not constitute human verification of every generated or rewritten
sentence.

\section{Reproducible Implementation}
\label{app:implementation}

\subsection{Architecture Specification}

The pretrained backbone is Qwen3-VL-8B-Instruct
\citep{bai2025qwen3vl}. The visual encoder and base language-model weights are
frozen. Table~\ref{tab:architecture_spec} lists the fixed dimensions of the
reported configuration. The Clip Event Adapter produces \(K_m=9\) clip-level event
tokens (one special token and eight slots). The active memory uses the same
token budget, and the memory-to-language projector produces \(K_p=8\) soft
prefix tokens.

\begin{table*}[t]
\centering
\small
\setlength{\tabcolsep}{5pt}
\renewcommand{\arraystretch}{1.07}
\begin{tabular}{@{}p{0.28\textwidth}p{0.22\textwidth}p{0.40\textwidth}@{}}
\toprule
Component / symbol & Reported value & Specification \\
\midrule
Backbone & Qwen3-VL-8B-Instruct & Frozen visual encoder and frozen base LLM \\
Visual/LLM hidden size \(d_v,d_l\) & 4096 / 4096 & Pretrained token spaces \\
Event-memory hidden size \(d_m\) & 1024 & Adapter, active memory, and completed event memory \\
Clip Event Adapter \(\mathcal A\) & 2 layers, 8 heads & Cross-attention resampler; \(K_m=9\) output tokens \\
Memory update \(\mathcal U\) & 2 layers, 8 heads & Memory-to-clip attention, self-attention, FFN, and per-token gate \(G_t\) \\
Memory initializer \(\mathcal I\) & 1 layer & Initializes a new active state from \(E_t\) \\
Memory-to-language projector \(P_\theta\) & 2 layers, 8 heads & \(K_p=8\) projected soft-prefix tokens \\
Maximum active duration \(D_{\max}\) & 24 s & Deterministic duration-rollover safeguard \\
LoRA for Stages 2--3 & \(r=8,\alpha=16\), dropout 0.05 & Targets \texttt{q\_proj}, \texttt{k\_proj}, \texttt{v\_proj}, and \texttt{o\_proj} \\
\bottomrule
\end{tabular}
\caption{Architecture specification for the paper-facing configuration.}
\label{tab:architecture_spec}
\end{table*}

The transition head compares the previous active state with the current clip
and receives action and age cues. On reset, the pre-reset state is exported as
\(M_j^{\mathrm c}\). Completed event memories enter \(\mathcal B_t\)
immediately, while record caption generation and deterministic record assembly
complete asynchronously before \(R_j\) becomes eligible for
\(\mathcal L_t\). Historical records are text supplied to the language prompt;
they are not additional latent memory tokens.

\subsection{Three-Stage Training}

Table~\ref{tab:training_spec} summarizes the configured training recipes. The
pretrained backbone tensors are loaded in bfloat16, and bfloat16 is enabled only
on supported trainer paths. The frozen artifacts do not establish automatic
mixed-precision execution for every trainer.

\begin{table*}[b]
\centering
\scriptsize
\setlength{\tabcolsep}{3.0pt}
\renewcommand{\arraystretch}{1.10}
\begin{tabular}{@{}p{0.12\textwidth}p{0.24\textwidth}p{0.15\textwidth}p{0.18\textwidth}p{0.22\textwidth}@{}}
\toprule
Stage & Trainable path & Per-process batch / accumulation & Learning rate and schedule & Main objective / budget \\
\midrule
1: streaming event memory
& Clip Event Adapter, event-memory encoder, transition and auxiliary event heads
& 1 / 16
& \(10^{-4}\); 500-step warmup; AdamW
& Teacher-forced reset; up to 50k steps; weights 0.5 clip action, 1.0 transition, 1.0 completed-event action set, 0.3 event type \\
2: memory-to-record
& Memory projector and language LoRA; base LLM frozen
& 1 / 16
& Projector \(10^{-4}\), LoRA \(10^{-5}\); 500-step warmup
& Assistant caption tokens only; projector alignment then LoRA warmup and joint fitting; target length 192 \\
3: multi-context commentary
& Language LoRA, memory projector, route/local/retrieval auxiliary heads; Stage-1 memory cached/frozen in the reported run
& 2 / 8 per process
& LoRA \(2\!\times\!10^{-5}\), projector \(3\!\times\!10^{-5}\); 500-step warmup
& Three-track response CE with auxiliary context losses; 10k configured budget; prompt/target/generation limits 2048/128/96 \\
\bottomrule
\end{tabular}
\caption{Three-stage training configuration. The displayed batch size is per
process. Stage~3 uses prompt profile \texttt{paper\_clean\_v2} and seed 42 for
the configured sampler.}
\label{tab:training_spec}
\end{table*}
\FloatBarrier

The memory-cache Stage~3 training path requires an explicit Stage~1 checkpoint
matching the materialized memory tokens and stores that frozen Stage~1 state in
the Stage~3 checkpoint. The local step-1,000 replay artifact does not retain the
training launch command or source-checkpoint hashes, so it does not independently
verify Stage~2 checkpoint initialization for this run. Stage~3 uses a
memory-cache curriculum: the first 5,000 configured steps use task-legal
teacher-forced context, while the next 5,000 expose runtime-selected context,
increasing the predicted-context ratio from 0.10 to 0.50 every 250 steps. The
main-text quality result uses checkpoint step 1,000 and is evaluated by
reference-anchored replay. This checkpoint lies in the teacher-forced phase and
has not undergone the second-phase runtime-context exposure. The frozen replay
protocol records the config, checkpoint, reference, and clip SHA-256 hashes
together with the effective inference settings.

\FloatBarrier
\subsection{Inference, Retrieval, and Scheduler Parameters}

Table~\ref{tab:runtime_params} separates memory/retrieval settings from the
rule-assisted speaking policy. The scheduler is deterministic and is included
to reproduce executed system behavior; it is not a learned policy and is not
ranked as a separate scientific contribution.

\begin{table}[t]
\centering
\footnotesize
\setlength{\tabcolsep}{3pt}
\begin{tabular}{@{}p{0.56\columnwidth}p{0.37\columnwidth}@{}}
\toprule
Runtime parameter & Value \\
\midrule
Clip duration / sampling & 4 s / 2 FPS \\
Local event-memory buffer horizon / maximum selected & 180 s / 4 events \\
Historical retrieval & embedding top-3 \\
Historical eligibility & same half, gap \(\ge 90\) s \\
Current-event cooldown & 12 s \\
Recent-window tick / minimum events & 120 s / 3 \\
Window cooldown after event output & 20 s \\
Historical cooldown / minimum records & 180 s / 8 \\
Historical score threshold & 0.12 \\
Scheduler priority & history \(>\) event \(>\) recent \\
Commentary decoding & deterministic, max 96 tokens \\
Record-caption decoding & deterministic, max 64 tokens \\
\bottomrule
\end{tabular}
\caption{Default runtime and rule-assisted scheduler settings. The replay in
Table~\ref{tab:common_anchor_quality} fixes output anchors and therefore does not evaluate these
scheduler thresholds.}
\label{tab:runtime_params}
\end{table}

Event closures enqueue caption jobs. The completed event memory is immediately
available, whereas an event record becomes historically retrievable only after
caption generation, deterministic assembly, and insertion complete. The
reference-anchored protocol deterministically drains pending jobs before
retrieval at a historical anchor;
the free-scheduler protocol does not. On the frozen main-table replay, eight
parallel workers process 86,145 clips and insert 20,689 ready records. These
workers partition independent half streams; state is not shared across halves.

\paragraph{Caption-queue and record-readiness timeline.}
At closure, the pre-reset state is exported as \(M_j^{\mathrm c}\), inserted
immediately into \(\mathcal B_t\), and enqueued as a caption job. It is not yet
part of the event-record LTM \(\mathcal L_t\). A drain projects the completed
memory, generates the event-record caption \(c_j\), attaches deterministic
identity, time, action, type, and match-state fields, and inserts the resulting
auditable record \(R_j\) when its generated caption is non-empty. Only a
non-empty record whose insertion has
completed is retrieval-eligible, so
\begin{align*}
M_j^{\mathrm c}\in\mathcal B_t
&\Longrightarrow \text{caption job}
\Longrightarrow c_j\ \text{ready},\\
c_j\ \text{ready}
&\Longrightarrow R_j\in\mathcal L_t.
\end{align*}
The cached-token replay batches at most four caption jobs. With its zero
timer-wait setting, normal draining occurs when a batch fills, before a pending
record would age out of the local event-memory buffer, or at stream end. A historical
reference anchor additionally forces pending jobs to drain, which is the
record-readiness intervention disclosed in Table~\ref{tab:protocol_matrix}.
Free-scheduler replay never drains because of a target anchor and therefore
observes natural asynchronous readiness. The frozen raw-video deployment runtime
interprets zero wait as immediate draining; its timing is reported
separately from cached replay.

\FloatBarrier
\subsection{Artifact Provenance}

Table~\ref{tab:artifact_map} provides stable labels for the results used in
Appendices A--D. These labels identify result-specific frozen sources without
exposing server absolute paths, usernames, or machine names.

\begin{table}[t]
\centering
\scriptsize
\setlength{\tabcolsep}{2.2pt}
\renewcommand{\arraystretch}{1.08}
\begin{tabular}{@{}p{0.18\columnwidth}p{0.31\columnwidth}p{0.18\columnwidth}p{0.24\columnwidth}@{}}
\toprule
ID & Result & Step / seed & Population \\
\midrule
\texttt{ANCHOR-S1K}
& Main-paper StreamSoccer commentary-quality row
& 1,000 / 42
& 63 matches; 3,789 anchors \\
\texttt{REP-MEM}
& Recurrent event memory representation row
& 3,000 / 42
& 49 matches; 1,726 events \\
\texttt{REP-EVENT}
& Operational-event pooling row
& 3,000 / 42
& 49 matches; 1,726 events \\
\texttt{REP-FIXED}
& Fixed-time pooling row
& 3,000 / 42
& 49 matches; 1,726 events \\
\bottomrule
\end{tabular}
\caption{Stable artifact labels for the protocol and representation results in
Appendices A--D.}
\label{tab:artifact_map}
\end{table}

Result-specific protocol and evaluation files, rather than this PDF, store the
available content hashes, commands, checkpoint identities, and scorer settings.
The frozen local records do not contain a unified release manifest or a code
revision hash, so this appendix does not claim either. The three
representation rows share the same
implementation, data order, targets, projector initialization, optimization
budget, and deterministic decoding settings; only the representation and its
fitted checkpoint differ.

\FloatBarrier
\section{Event-Memory Mechanism Evidence}
\label{app:event_memory_evidence}

\subsection{Matched Intermediate-Representation Comparison}

We directly compare three visual summaries for event-record caption generation.
Recurrent event memory exports \(M_j^{\mathrm c}\) when an operational event
closes. Operational-event pooling averages clip-level event tokens over the
same event support, isolating the value of recurrent state formation. Fixed-time
pooling averages at most six clips from the trailing 24 s, providing a simple
temporal-window control. Every row supplies a \(9\times1024\) representation to
an independently fitted but architecturally identical projector and language
readout initialized from the same state.

\begin{table}[t]
\centering
\scriptsize
\setlength{\tabcolsep}{2.0pt}
\renewcommand{\arraystretch}{1.06}
\begin{tabular}{@{}p{0.26\columnwidth}ccccc@{}}
\toprule
Representation & Tok-F1 & B@4 & METEOR & ROUGE-L & CIDEr \\
\midrule
Recurrent memory
& \textbf{0.2642} & \textbf{0.2777} & \textbf{0.2358} & \textbf{0.4335} & \textbf{37.40} \\
Event pooling
& 0.2290 & 0.2556 & 0.2192 & 0.3986 & 22.20 \\
Fixed-time pooling
& 0.2108 & 0.2462 & 0.2104 & 0.3916 & 15.72 \\
\bottomrule
\end{tabular}
\caption{Matched memory-to-record representation comparison on 1,726 events
from 49 matches. CIDEr uses the \(100\times\) scale.}
\label{tab:representation_extended}
\end{table}

\begin{table}[t]
\centering
\footnotesize
\setlength{\tabcolsep}{3pt}
\begin{tabular}{@{}lcc@{}}
\toprule
Tok-F1 difference & Mean & 95\% paired CI \\
\midrule
Memory \(-\) event pooling & 0.0351 & [0.0259, 0.0444] \\
Memory \(-\) fixed pooling & 0.0533 & [0.0448, 0.0625] \\
Event pooling \(-\) fixed pooling & 0.0182 & [0.0114, 0.0252] \\
\bottomrule
\end{tabular}
\caption{Tok-F1 differences from 10,000 paired match-cluster bootstrap
replicates.}
\label{tab:representation_bootstrap}
\end{table}

All three methods are trained for 3,000 optimizer steps with global batch 16,
giving 48,000 sampled-event exposures per row. Inference is deterministic with
a 64-token generation limit. Recurrent memory exceeds operational-event
pooling by 0.0351 Tok-F1 and fixed-time pooling by 0.0533; the paired intervals
exclude zero. Operational-event pooling also exceeds fixed-time pooling by
0.0182 Tok-F1, with 95\% interval [0.0114, 0.0252]. The
event-pooling--versus--fixed-time contrast supports event-aligned temporal
support, whereas the recurrent-memory--versus--event-pooling contrast supports
learned state formation under matched event support.

This experiment is deliberately bounded. It uses one seed and the legacy
5,676-train/1,726-test protocol. The final checkpoints are compared at a fixed
step rather than selected independently. The 1,726-event population is a
legacy development/evaluation population rather than a never-observed test
set. The stored decoder drops some
leading special-token fragments. A deterministic text-only sensitivity repair
yields Tok-F1 values of 0.3093, 0.2705, and 0.2566 for recurrent memory, event
pooling, and fixed-time pooling, respectively, and preserves the three-method
ordering. Table~\ref{tab:representation_extended} reports the unrepaired frozen
decoder outputs and supports matched-budget representation decodability only;
it does not establish never-observed-test generalization, record factuality, or
multi-seed training stability.

\section{Commentary Quality, Memory Scope, and Retrieval}
\label{app:commentary_quality}

\subsection{Expanded Commentary-Quality Results}

Table~\ref{tab:streamsoccer_extended_quality} adds metrics and coverage for the
StreamSoccer row in the main commentary-quality table. These values come from
the same step-1,000 reference-anchored replay described in
Appendix~\ref{app:evidence_contract}; they do not constitute a separate
free-scheduler evaluation. The replay emits one valid output at every one of
the 3,789 test anchors, with no empty generations.

\begin{table*}[t]
\centering
\small
\setlength{\tabcolsep}{4.0pt}
\renewcommand{\arraystretch}{1.06}
\begin{tabular}{@{}lrrrrrrrr@{}}
\toprule
Track & \(N\) & Valid & Tok-F1 & B@4 & METEOR & ROUGE-L & CIDEr & BS \\
\midrule
Current-event
& 2,050 & 100\% & 0.3234 & 0.2929 & 0.2487 & 0.4606 & 38.62 & 0.9734 \\
Recent-window
& 1,009 & 100\% & 0.3499 & 0.0652 & 0.1518 & 0.2680 & 23.96 & 0.9612 \\
Historical-memory
& 730 & 100\% & 0.2292 & 0.0451 & 0.1441 & 0.2146 & 17.39 & 0.9636 \\
\bottomrule
\end{tabular}
\caption{Additional StreamSoccer commentary metrics under reference-anchored
replay. CIDEr uses the \(100\times\) scale, BS denotes BERTScore-F1, and valid
coverage is the fraction of requested anchors producing a non-empty output.}
\label{tab:streamsoccer_extended_quality}
\end{table*}

Retrieval traces provide a separate diagnostic of historical-record
availability. Among 324 frozen retrieval-loss queries with positive
supervision, embedding retrieval obtains
R@1 \(=0.0741\), R@3 \(=\) R@5 \(=0.1327\), and MRR \(=0.1003\).
In total, 371 anchored outputs contain retrieved records, and 43.92\% of
scheduler decisions are retrieval-ready under this controlled replay. These
figures characterize the frozen retriever and do not establish that every
retrieved record improves the generated commentary.

\subsection{Blinded LLM Semantic Audit}
\label{app:blind_llm_audit}

We conduct a fixed-protocol four-way blinded audit over all 3,789 test anchors from 63
matches. For each anchor, a temperature-zero GPT-5.6-terra judge receives the
silver reference, task description, and four candidate slots under a
sample-specific deterministic permutation. Model identities are withheld from
the prompt. A missing model output remains in its randomized slot as
\texttt{NO\_RESPONSE} and cannot win; every strict win rate therefore uses the
full 3,789-anchor population rather than a valid-output-only subset. All four
models respond on 1,086 anchors, and at least one external model has
\texttt{NO\_RESPONSE} on the remaining 2,703 anchors. Response rate is reported
beside preference so availability is visible alongside preference.

\begin{table*}[t]
\centering
\scriptsize
\setlength{\tabcolsep}{3.3pt}
\renewcommand{\arraystretch}{1.08}
\begin{tabular}{@{}lrrrrr@{}}
\toprule
Method & Current & Recent & Historical & Overall [95\% CI] & Response \\
\midrule
StreamSoccer
& \textbf{90.73} & \textbf{96.63} & \textbf{95.21}
& \textbf{93.16 [92.32, 94.01]} & 100.00 \\
StreamingVLM
& 8.68 & 3.37 & 4.38 & 6.44 [5.65, 7.21] & 93.77 \\
TimeChat-Online
& 0.59 & 0.00 & 0.41 & 0.40 [0.21, 0.61] & 41.20 \\
VideoLLM-online
& 0.00 & 0.00 & 0.00 & 0.00 [0.00, 0.00] & 64.61 \\
\bottomrule
\end{tabular}
\caption{Strict win rate and response rate (\%) in the frozen blinded LLM
semantic audit. Current, recent, and historical contain 2,050, 1,009, and 730
anchors. Overall intervals use 10,000 match-cluster bootstrap replicates over
63 matches; missing outputs remain in the fixed denominator and cannot win.}
\label{tab:blind_llm_semantic_audit}
\end{table*}

The judge selects StreamSoccer on 3,530 of 3,789 anchors, giving a strict win
rate of 93.16\% with match-clustered 95\% CI [92.32\%, 94.01\%]. This ordering
holds on each commentary track. The randomized label positions A--D receive
960, 971, 911, and 947 wins, respectively (25.34\%, 25.63\%, 24.04\%, and
24.99\% of judged anchors), so the result is not concentrated in one displayed
position. Under the declared \texttt{NO\_RESPONSE}-cannot-win protocol, StreamSoccer is the
preferred available response against the silver reference across the fixed
test population. This single-judge audit does not convert silver references
into human preference labels or establish claim-level factual correctness. It
ranks final outputs and does not determine whether a retrieved historical
record supports or causes a historical-memory response.

\subsection{Memory-Scope Analysis}

The main text compares separately trained configurations with progressively
broader accessible context. Table~\ref{tab:memory_scope_cider} reports the
locally verified CIDEr values alongside the protocol details in this
appendix. These rows are configuration comparisons: changing memory scope
also changes the context and objectives available during training, so the
differences are not inference-only causal ablations of one checkpoint.

\begin{table}[t]
\centering
\small
\setlength{\tabcolsep}{3.5pt}
\renewcommand{\arraystretch}{1.06}
\begin{tabular}{@{}lrrr@{}}
\toprule
Configuration & Current & Recent & Historical \\
\midrule
Current memory only & 33.03 & 15.06 & 13.22 \\
\(+\) local event-memory buffer & 38.57 & 17.96 & 17.22 \\
Full three-scope model & \textbf{38.62} & \textbf{23.96} & \textbf{17.39} \\
\bottomrule
\end{tabular}
\caption{Verified CIDEr (\(100\times\)) results for the separately trained
memory-scope configurations.}
\label{tab:memory_scope_cider}
\end{table}

Each frozen evaluation summary contains 3,789 exact-ID matches and no unmatched
outputs. The current-only and local event-memory buffer rows use their respective
step-10,000 retrained checkpoints; the full row uses the selected step-1,000
full-model checkpoint. Relative to the current-only configuration, the
local event-memory buffer configuration is higher by 5.54, 2.89, and 4.00 CIDEr on the
current, recent, and historical tracks. Relative to the local event-memory buffer
configuration, the full configuration is higher by 0.05, 6.01, and 0.17 CIDEr,
respectively. These are configuration-level differences and do not isolate the
causal effect of retrieved historical records on a particular generation.

\subsection{Historical-Record Intervention}

The retrieval diagnostics above measure whether the expected source is found,
but they do not isolate whether the language model uses a relevant record.
We therefore perform a controlled intervention on historical-positive anchors
while holding the checkpoint, current and local context, output anchor, and
decoding settings fixed. The three non-empty arms each use three records,
whereas the no-record arm removes the historical block.

Table~\ref{tab:historical_record_intervention} reports the completed
same-checkpoint intervention on a frozen set of 209 historical-memory anchors
from 56 matches (79 half streams). Current/local latent context, output anchor,
decoding, and the non-historical prompt are identical across arms; only the
textual historical-record block changes. Every requested output is non-empty.
The runtime-ranked arm bypasses the learned NULL decision solely to impose the
same three-record input budget as the other non-empty arms; it is therefore a
controlled retrieval diagnostic rather than the native online policy.

\begin{table}[t]
\centering
\scriptsize
\setlength{\tabcolsep}{3.8pt}
\renewcommand{\arraystretch}{1.06}
\resizebox{\columnwidth}{!}{%
\begin{tabular}{@{}lrrrrr@{}}
\toprule
Historical input & Records & Tok-F1 & B@4 & CIDEr & BS \\
\midrule
No historical record & 0 & 0.1426 & 0.0169 & 2.61 & 0.8543 \\
Matched hard-negative top-3 & 3 & 0.2724 & 0.0726 & 30.92 & 0.8922 \\
Runtime-ranked top-3$^{\dagger}$ & 3 & 0.2682 & 0.0697 & 30.15 & 0.8906 \\
Oracle-ID top-3$^{\ddagger}$ & 3 & 0.2819 & 0.0841 & 37.96 & 0.8947 \\
\bottomrule
\end{tabular}
}
\caption{Automatic overlap diagnostics for the four-arm historical-record
intervention. All arms use the identical 209-anchor population and have 100\%
valid coverage. CIDEr uses the \(100\times\) scale and BS denotes
BERTScore-F1. $^{\dagger}$The learned NULL decision is bypassed only to fix the
three-record budget. $^{\ddagger}$The positive record is placed at rank 1;
this is a controlled oracle, not native online performance.}
\label{tab:historical_record_intervention}
\end{table}

Injecting matched hard-negative records instead of no historical record raises
per-anchor token-F1 by 0.1298 (match-clustered 95\% CI
[0.1157, 0.1441]). Oracle-ID injection raises it by 0.1393 over no historical
record (95\% CI [0.1253, 0.1538]). The paired intervals for runtime minus matched
hard negatives ($\Delta=-0.0042$; 95\% CI [-0.0222, 0.0137]) and oracle minus
runtime ($\Delta=+0.0137$; 95\% CI [-0.0043, 0.0318]) both include zero. Among
the three non-empty arms with a fixed top-3 budget, the audit detects no
automatic-metric advantage of runtime-ranked records over hard negatives and
no automatic-metric gap to the oracle arm. The no-record contrasts show that
adding a historical-record block changes the generator's automatic overlap,
but neither comparison establishes semantic use of relevant historical
evidence or superior runtime retrieval.

\FloatBarrier
\section{Continuous Execution and Long-History Efficiency}
\label{app:continuous_execution}

\subsection{Raw-Video Efficiency}

The raw-video benchmark includes video decoding, online visual encoding,
persistent state updates, and the fixed evaluation workload. It is separate
from the native free-scheduler policy. Table~\ref{tab:raw_video_efficiency}
reports p95 cumulative real-time factor (RTF) and the p95 peak VRAM at the last
supported horizon.

\begin{table*}[t]
\centering
\scriptsize
\setlength{\tabcolsep}{4.8pt}
\renewcommand{\arraystretch}{1.08}
\begin{tabular}{@{}lrrrrr@{}}
\toprule
Method & T15 & T30 & T45 & T90 & Peak VRAM \\
\midrule
StreamingVLM
& 0.274 & 0.285 & 0.291 & 0.285 & 17,072 MiB \\
TimeChat-Online\textsuperscript{*}
& 0.138 & 0.231 & 0.322 & -- & 77,603 MiB \\
VideoLLM-online
& 0.171 & 0.259 & 0.328 & 0.559 & 50,969 MiB \\
StreamSoccer
& \textbf{0.110} & \textbf{0.119} & \textbf{0.119} & \textbf{0.122}
& 17,996 MiB \\
\bottomrule
\end{tabular}
\caption{Raw-video efficiency. T15--T90 are p95 cumulative
RTFs after 15, 30, 45, and 90 minutes of observed match time. Peak VRAM is the
p95 value at the last valid horizon. \textsuperscript{*}TimeChat-Online has no
T90 value because its supported context ends before that horizon.}
\label{tab:raw_video_efficiency}
\end{table*}

StreamSoccer's source-clean cohort contains 58 matches and 174 complete runs.
The full-match RTF has median 0.1163 and p95 0.1221; a match-cluster bootstrap
gives a median-RTF interval of
[0.1137, 0.1183]. Whole-run peak VRAM has median 17,992 MiB and p95
17,996 MiB.

\subsection{Record Readiness and Latency Breakdown}

Table~\ref{tab:stage_wall_breakdown} decomposes the measured StreamSoccer
workload into auditable wall-time components. The entries are component-wise
p95 RTFs computed over the full run; because their tails need not occur on the
same request, their sum is not an isolated commentary latency.

\begin{table}[t]
\centering
\small
\setlength{\tabcolsep}{5pt}
\begin{tabular}{@{}lr@{}}
\toprule
Component & p95 RTF \\
\midrule
Raw video decode & 0.0053 \\
Visual encoding & 0.0273 \\
State update & 0.0040 \\
External generation wall & 0.0201 \\
Native generation wall & 0.0692 \\
\bottomrule
\end{tabular}
\caption{Diagnostic component-wise p95 RTFs for the frozen raw-video
StreamSoccer workload. Prompt construction, post-processing, and residual
overhead are not separately tabulated.}
\label{tab:stage_wall_breakdown}
\end{table}

Request-level end-to-end p95 latency is 2.467, 3.826, 3.855, and 3.801 s at
T15, T30, T45, and T90, respectively. Record-caption work is asynchronous:
completed event memory enters the local event-memory buffer immediately, but a historical
record becomes eligible only after captioning, assembly, and insertion.

To quantify the reference-triggered readiness intervention, we audit the 730 historical-memory
anchors against an otherwise identical natural-queue shadow that disables
anchor-triggered draining. Table~\ref{tab:record_readiness_audit} reports the
paired result. Reference anchors trigger pending caption consolidation at 594
anchors and drain 1,269 caption jobs. Every force-drained job corresponds to
an event ending 4--76~s before its anchor, below the frozen 90-s historical
eligibility gap. Consequently, reference-induced consolidation changes neither
the eligible candidate set nor retrieved record identities, and it makes no
positive historical evidence newly available at any audited anchor.

\begin{table}[t]
\centering
\small
\setlength{\tabcolsep}{4pt}
\begin{tabular}{@{}lr@{}}
\toprule
Audit item & Value \\
\midrule
Historical-memory anchors & 730 \\
Anchors with pending jobs (forced / shadow) & 594 / 603 \\
Reference-triggered drain batches & 594 \\
Unique force-drained caption jobs & 1,269 \\
Eligible candidate set changed & 0 / 730 \\
Retrieved record identities changed & 0 / 730 \\
Positive-evidence readiness changed & 0 / 730 \\
\bottomrule
\end{tabular}
\caption{Paired record-readiness audit of the 730 historical-memory anchors.
The forced condition reproduces reference-anchored replay; the natural-queue
shadow disables only anchor-triggered caption draining.}
\label{tab:record_readiness_audit}
\end{table}

\subsection{Free-Scheduler Execution Audit}

We additionally audit a frozen free-scheduler replay to demonstrate that the
state, caption queue, retrieval, and rule-assisted speaking path execute over
complete half streams. This run uses checkpoint step 7,000, cached Qwen tokens,
predicted operational transitions, embedding retrieval with same-half and
90-s-gap eligibility, and label-assisted action/type fields. It is an execution
audit rather than a ranking of scheduler strategies.

\begin{table}[t]
\centering
\footnotesize
\setlength{\tabcolsep}{3.5pt}
\begin{tabular}{@{}lr@{}}
\toprule
Audit item & Value \\
\midrule
Half streams / clips & 126 / 86,145 \\
Commentary outputs & 5,949 \\
Comments per minute & 1.0505 \\
Current / recent / historical & 2,976 / 2,820 / 153 \\
Caption jobs & 20,658 \\
Ready / empty captions & 20,654 / 4 \\
Exact / near-duplicate rate & 0.000168 / 0 \\
Event-to-comment delay & 4.0 s \\
Total latency p50 / p95 & 2.357 / 4.209 s \\
Scheduler steps with retrieval-ready records & 82,770 / 86,145 (96.08\%) \\
\bottomrule
\end{tabular}
\caption{Continuous free-scheduler execution audit. The scheduler is
rule-assisted and the run retains label-assisted action/type metadata.}
\label{tab:free_scheduler_audit}
\end{table}

Only 88 of 5,949 free-scheduler outputs occur within one second of a reference
anchor. Text metrics on that small matched subset would therefore confound
coverage and timing with language quality and are not used as commentary
quality evidence.

\section{Auditable Cases and Failure Analysis}
\label{app:cases_failures}

\subsection{Case-Selection Protocol}

The gallery uses only frozen test outputs. For each of the three commentary
tracks, we retain the five pre-indexed candidates in the frozen case manifest;
row \(i\) aligns candidate \(i\) from the current-event, recent-window, and
historical-memory tracks. This fixed five-by-three ordering is recorded with
the sample IDs in the frozen selection manifest and is not changed after rendering.
Each card shows causal visual evidence, the reference, three external streaming
VLM outputs, and the StreamSoccer output. The displayed text is a literal,
compact prefix of the frozen output; the full output and exact color-span audit
remain in the associated case-card artifact.

\subsection{Success and Failure Gallery}

Figure~\ref{fig:supp_three_track_cases} lays out five aligned rows of examples.
Columns correspond to current-event, recent-window, and historical-memory
commentary, respectively; each card retains only visual evidence frames, with
no timestamps or timeline. Light blue-grey identifies the reference, amber
identifies external streaming VLMs, and blue identifies StreamSoccer. Red text
marks wording heuristically flagged as unsupported, repetitive, malformed, or
generic; green text marks StreamSoccer wording heuristically aligned to the
reference event/evidence. These colors are qualitative reading aids rather than
population-level error rates.

\begin{figure*}[t]
\centering
\includegraphics[width=0.97\textwidth]{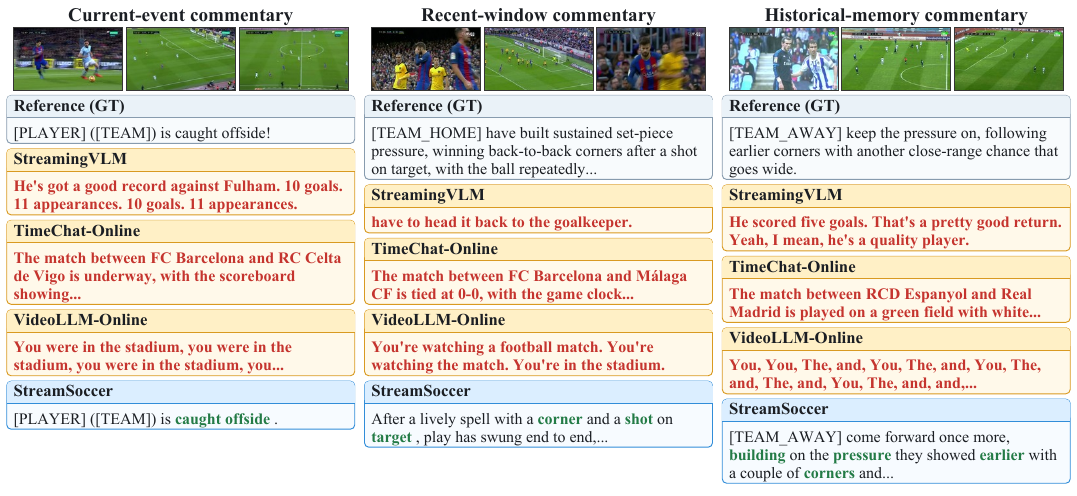}
\includegraphics[width=0.97\textwidth]{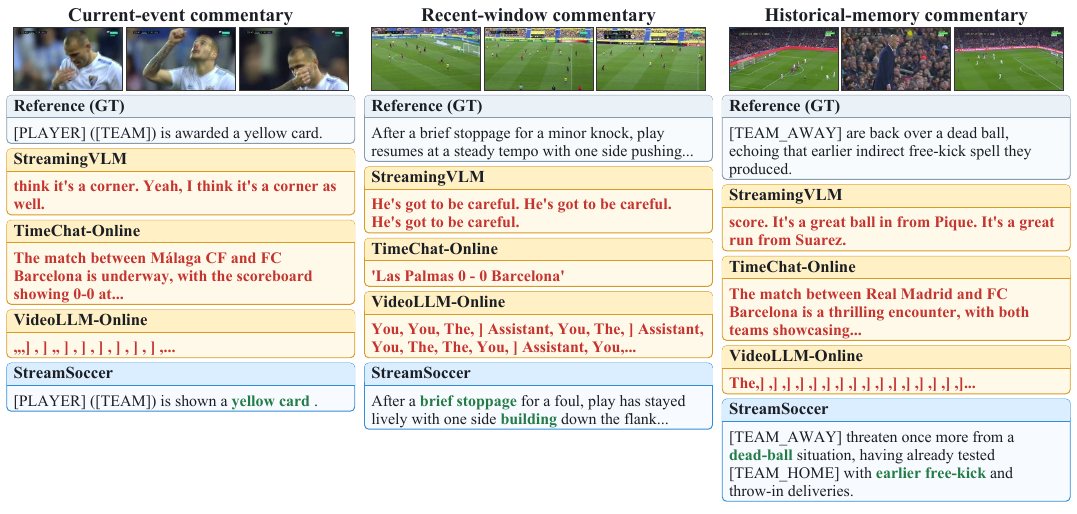}
\includegraphics[width=0.73\textwidth]{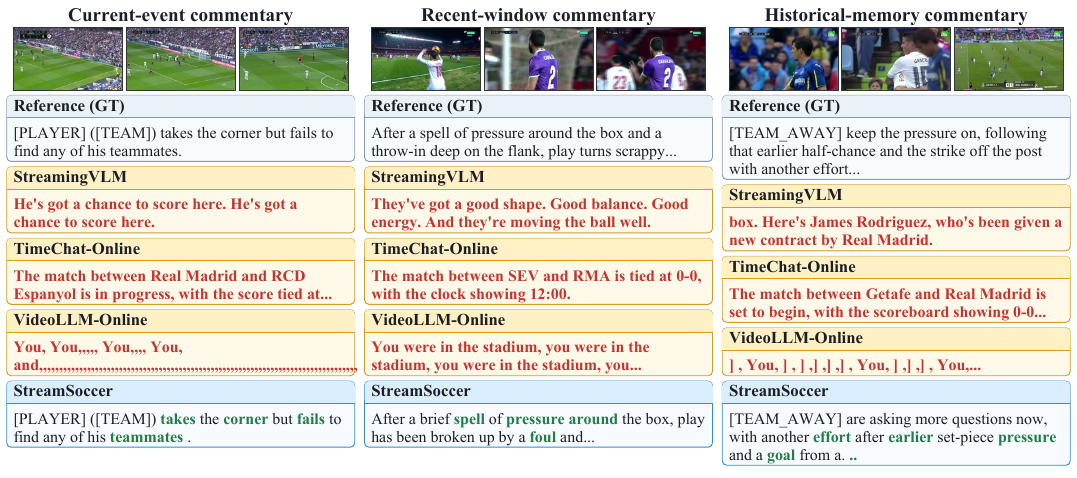}
\caption{Five qualitative rows, each containing one frozen case from the
current-event, recent-window, and historical-memory commentary tracks. The
three column labels define the task type; no event timestamps or timelines are
shown. Full frozen outputs, sample IDs, and coloring reasons are retained in
the accompanying case-card audit. Rows 1--3 are shown here and the figure
continues on the following pages.}
\label{fig:supp_three_track_cases}
\end{figure*}

\begin{figure*}[t]
\ContinuedFloat
\centering
\includegraphics[width=0.985\textwidth]{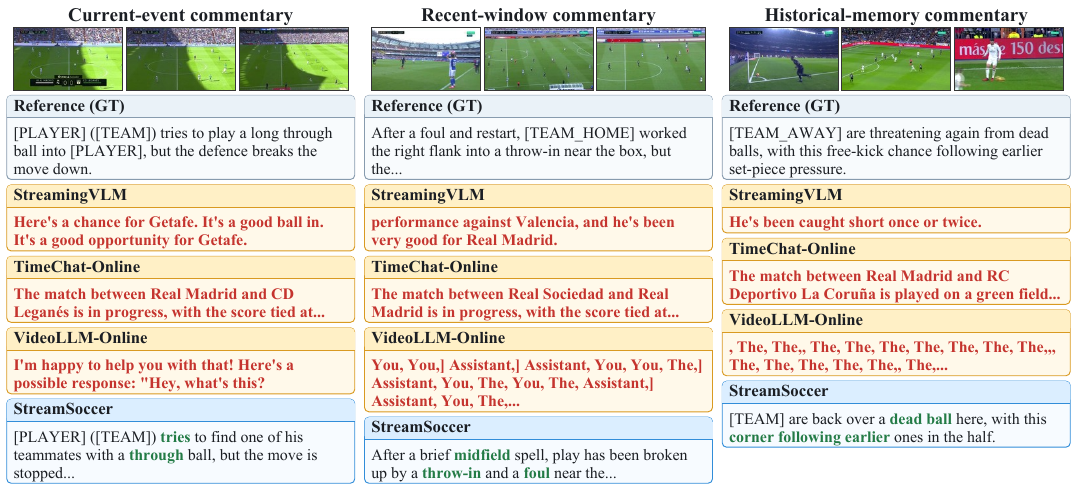}
\includegraphics[width=0.985\textwidth]{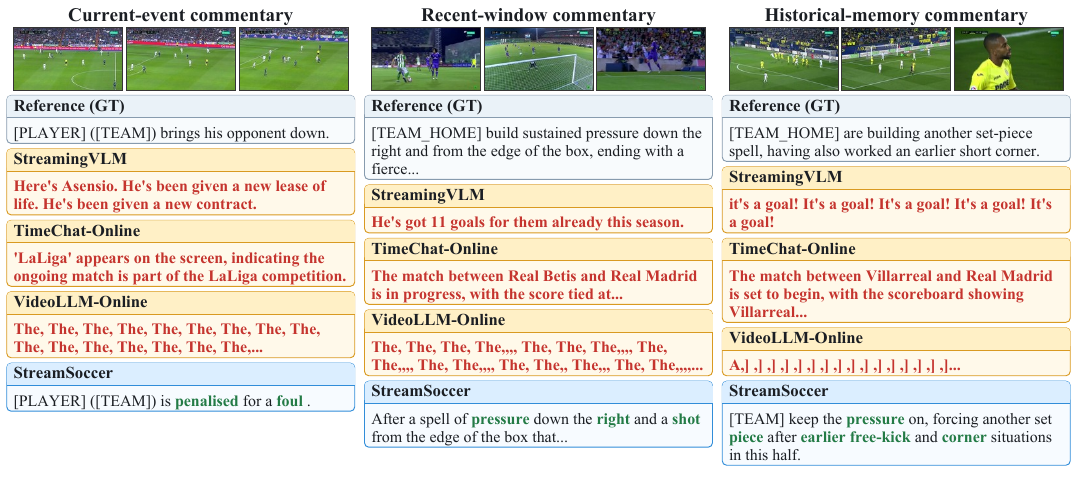}
\caption[]{Qualitative gallery, continued (rows 4--5).}
\end{figure*}
\FloatBarrier

\subsection{Mechanically Audited Failures}

This subsection reports deterministic mechanical failures only. The fixed-protocol
four-way audit in Appendix~\ref{app:blind_llm_audit} ranks the four permuted
candidate slots, including \texttt{NO\_RESPONSE}, and does not assign
claim-level semantic-error labels; no population-level semantic-error
frequency is reported. In the step-10,000 reference-anchored replay, 4 of
3,789 commentary outputs contain CJK or mojibake characters, including one
output containing both repetition and encoding contamination. The same audit
flags 38 of 20,689 generated caption records. In the step-7,000 free-scheduler replay, 4 of
20,658 record captions are empty; the commentary exact-duplicate rate is
0.000168 and the near-duplicate rate is zero.

\section{Scope, Limitations, and Artifact Index}
\label{app:scope_artifacts}

\subsection{Scope and Limitations}

The following boundaries apply to the evidence in the main text and this
appendix.

\begin{itemize}
\item Operational events are deterministic, rule-derived units for state
control, not unconstrained discovery of natural semantic events. The 24-s
rollover safeguard closes 40.2\% of constructed events.
\item MatchTime alignment covers 9.13\% (15,189 / 166,419) of all operational events and
therefore emphasizes moments for which commentary supervision is available.
\item The main commentary table is reference-anchored and uses oracle
operational closure, current-event alignment, label-assisted action/type
metadata, and a record-readiness intervention.
\item The free-scheduler audit uses checkpoint step 7,000, cached visual tokens,
and label-assisted action/type metadata. It demonstrates continuous execution,
not a purely visual, target-independent commentary benchmark.
\item The raw-video benchmark uses a fixed current-event/recent-window/historical-memory request
workload to compare history scaling; it does not evaluate each method's native
speaking decisions.
\item The deployed scheduler is deterministic and rule-assisted. Scheduler
strategy learning is outside the paper's central claim.
\item Generated event-record captions are model outputs rather than verified
facts. Record availability is evaluated separately from downstream commentary
quality, and no record-level factual-correctness claim is made.
\item The blinded semantic audit uses one temperature-zero LLM judge and silver
references. It measures availability-aware, reference-conditioned preference
under the frozen prompt rather than human preference or claim-level factuality.
\item Experiments use one soccer data ecosystem. Cross-league,
cross-language, and broadcast-style generalization have not been established.
\item Retained context-token count is not instrumented in the current raw-video
artifact. Zero-valued context fields in diagnostic logs must not be interpreted
as proof that no context was retained.
\end{itemize}

\subsection{Reproduction Map}

Table~\ref{tab:extended_artifact_map} extends the stable labels in
Table~\ref{tab:artifact_map} to every frozen result and audit reported in this
appendix.

\begin{table}[t]
\centering
\footnotesize
\setlength{\tabcolsep}{3pt}
\renewcommand{\arraystretch}{1.03}
\begin{tabular}{@{}p{0.25\columnwidth}p{0.69\columnwidth}@{}}
\toprule
Artifact ID & Paper-facing role and frozen evidence records \\
\midrule
\texttt{ANCHOR-S1K}
& Reference-anchored StreamSoccer quality and retrieval diagnostics.
\textit{Records:} protocol and content hashes; command; traces; predictions;
evaluation summary \\
\texttt{REP-MEM}
& Recurrent event-memory representation result. \textit{Records:} checkpoint
hash; predictions; scores; paired bootstrap \\
\texttt{REP-EVENT}
& Operational-event pooling result. \textit{Records:} checkpoint hash;
predictions; scores; paired bootstrap \\
\texttt{REP-FIXED}
& Fixed-time pooling result. \textit{Records:} checkpoint hash; predictions;
scores; paired bootstrap \\
\texttt{SCOPE-CUR}
& Current-memory-only configuration. \textit{Records:} replay command and
evaluation summary \\
\texttt{SCOPE-LOCAL}
& Current plus local event-memory buffer configuration. \textit{Records:}
replay command and evaluation summary \\
\texttt{SCOPE-FULL}
& Full three-scope configuration. \textit{Records:} replay command and
evaluation summary \\
\bottomrule
\end{tabular}
\caption{Reproduction map for the frozen results and audits. Each stable ID
names the evidence records supporting the corresponding paper-facing result.}
\label{tab:extended_artifact_map}
\end{table}

\begin{table}[t]
\ContinuedFloat
\centering
\footnotesize
\setlength{\tabcolsep}{3pt}
\renewcommand{\arraystretch}{1.03}
\begin{tabular}{@{}p{0.25\columnwidth}p{0.69\columnwidth}@{}}
\toprule
Artifact ID & Paper-facing role and frozen evidence records \\
\midrule
\texttt{RAW-EWL}
& Raw-video efficiency, coverage, and stage timing. \textit{Records:} run
manifest; per-request timings; aggregate tables \\
\texttt{FREE-S7K}
& Free-scheduler continuous-execution audit. \textit{Records:} traces; queue
audit; outputs; summary \\
\texttt{HIST-CF209}
& Four-arm historical-record intervention. \textit{Records:} protocol;
209-anchor manifest; 836 outputs; evaluation summary; match-cluster bootstrap \\
\texttt{READY-730}
& Paired record-readiness audit. \textit{Records:} protocol; 730 paired rows;
drain traces; audit summary \\
\texttt{BLIND-3789}
& Four-way blinded LLM semantic audit. \textit{Records:} protocol and source
hashes; 3,789 permuted ballots; decoded judgments; match-cluster bootstrap \\
\texttt{FAIL-S10K}
& Step-10,000 reference-anchored mechanical failure audit. \textit{Records:}
commentary and record outputs; character, repetition, and encoding audit;
aggregate counts \\
\texttt{CASE-FROZEN}
& Fixed qualitative gallery. \textit{Records:} selection manifest; displayed
records; exact outputs; highlight/render audit \\
\bottomrule
\end{tabular}
\caption[]{Reproduction map, continued.}
\end{table}

The map uses stable IDs instead of absolute machine paths. Each result is backed
by a frozen run-specific protocol or manifest together with the listed
row-level outputs or traces and aggregate summary. Large row-level files are
not duplicated in this PDF. The historical intervention, record-readiness
audit, blinded semantic audit, and case-gallery selection are complete.

\balance
\section*{LLM Usage}
ChatGPT was used to aid in polishing the writing, including grammar,
readability, and sentence clarity. Separately, GPT-5.6-terra was used as the
single evaluator in the blinded semantic audit in
Appendix~\ref{app:blind_llm_audit}. It ranked anonymized candidate commentaries
against a silver reference under a frozen prompt; its judgments were used only
for evaluation and were not used as training targets.

\end{document}